\documentclass[11pt,a4paper]{article}

\usepackage{amsmath,amssymb,bm}
\usepackage{booktabs}
\usepackage{tabularx}
\usepackage{multirow}
\usepackage{array}
\usepackage{graphicx}
\usepackage{float}
\usepackage{caption}
\usepackage[section]{placeins}
\usepackage{xcolor}
\usepackage[margin=2.2cm]{geometry}
\usepackage{hyperref}
\usepackage{cite}
\hypersetup{colorlinks=true,linkcolor=blue,urlcolor=blue}

\title{AI4Plasma II: Geometry-Aware Surrogate Modeling for Variable Electrode Gap GEC Plasmas}
\date{}

\long\def\authorblock{%
\parbox{0.95\textwidth}{\centering
Hao~Yang$^{1,2}$, Zhenyu~Zou$^{1,3}$, Wenshuai~Jie$^{4}$, Luowei~Yin$^{1,6}$,
Yufeng~Wei$^{1,*}$, Daniel~Zhengyu~Huang$^{5,*}$, Chun~Chen$^{3}$,
Kangxin~Liu$^{1}$, Rubing~Han$^{1}$, Haoran~Guan$^{1}$, Yipei~Chen$^{1}$,
Peiting~You$^{1}$, and Yuwei~Fan$^{1}$\par\vspace{0.8em}
{\small
$^{1}$Theory Lab, Huawei Leibniz Research Center, Shatin, Hong Kong SAR, People's Republic of China\par
$^{2}$School of Aeronautic Science and Engineering, Beihang University, Beijing 100191, China\par
$^{3}$Department of Mechanical and Automation Engineering, The Chinese University of Hong Kong, Shatin, N.T. 999077, Hong Kong SAR, China\par
$^{4}$School of Mathematical Sciences, Peking University, Beijing 100871, China\par
$^{5}$Beijing International Center for Mathematical Research, Peking University, Beijing 100871, China\par
$^{6}$Department of Mathematics, Hong Kong University of Science and Technology, Clear Water Bay, Kowloon, Hong Kong SAR, China\par\vspace{0.4em}
$^{*}$Corresponding authors: Yufeng~Wei
(\href{mailto:wei.yufeng@huawei.com}{wei.yufeng@huawei.com}) and Daniel~Zhengyu~Huang
(\href{mailto:huangdz@bicmr.pku.edu.cn}{huangdz@bicmr.pku.edu.cn})}
}}
\author{\authorblock}

\begin{document}
\maketitle
\begin{abstract}
Industrial plasma process development requires broad exploration of operating and geometric parameters. Geometric parameters, particularly electrode spacing, affect discharge structure and processing uniformity. Exploring these parameter spaces through repeated high-fidelity simulations can be computationally expensive. Surrogate models can offer rapid predictions, but geometric variation complicates their application because the computational domain and discretization may change across samples.
In this work, six geometry-aware surrogate models based on prescribed and learned geometry mappings, extended-domain representation, continuous coordinates, and geometry-native discretizations are compared in terms of their data requirements for accurate predictions of full plasma fields and ion density profiles near the wafer. 
The simulation dataset spans pressure, voltage, frequency, and electrode gap, with each reference case requiring an average simulation cost of $5.32$ CPU h. 
To achieve a $2\%$ mean error across the four plasma fields, prescribed-mapping Geo-FNO and Transolver require 100 training cases, compared with 250--1000 for the other models. Predicting ion density profiles near the wafer is more demanding and reveals larger differences among models, with prescribed-mapping Geo-FNO requiring the fewest training cases for the profile accuracy targets considered. To further reduce the data required for accurate predictions near the wafer, we introduce charge-separation weighting into the training loss, which reduces the prescribed-mapping Geo-FNO's minimum tested training requirement for a $5\%$ profile error from 1000 to 100 cases.
These results support prescribed-mapping Geo-FNO as a practical starting point for the variable-gap problem studied here when a suitable geometry mapping is available, while Transolver achieves the highest prediction accuracy with sufficient training data. 

\end{abstract}

\section{Introduction}

Modern semiconductor fabrication is characterized by increasingly
fine features and geometrically complex device architectures, which demand
precise and uniform processing across the
wafer~\cite{kanarik2023human,ko2023optimization,oehrlein2024future,kanarik2015overview}.
Capacitively coupled plasmas (CCPs) play a central role in semiconductor plasma
etching, where ions accelerated through the sheath enable the anisotropic
transfer of nanoscale patterns~\cite{coburn1979ion,donnelly2013etching}.
Achieving the desired processing rate and uniformity requires control of the
plasma conditions near the wafer, which depend on both operating parameters and reactor
geometry~\cite{wang2021heating,lieberman2002standing}. Geometric parameters such
as electrode spacing and wafer-edge ring geometry can modify both the magnitude
and spatial distribution of plasma density and ion fluxes to the
wafer~\cite{liang2015uniformity,ahn2025sheath,kim2021nonuniformity,wang2025uniformity}.
Experiments and two-dimensional simulations have shown that electrode spacing
affects radial plasma uniformity and discharge behavior near the electrode
edge~\cite{steffens2003gap,zhou2022chamber,liang2015uniformity}. These effects
make electrode spacing an important variable in reactor design and highlight
the need for accurate and efficient prediction of plasma responses across
different gaps.

Physics based simulations provide spatially and temporally resolved descriptions
of discharge dynamics within the reactor. Fluid models describe charged
particle and energy transport through moment equations, kinetic models resolve
particle distribution functions, and hybrid models combine different
descriptions for selected plasma processes~\cite{vandijk2009plasma,donko2011particle,
sommerer1992hybrid,kushner2009hybrid,alves2018foundations}. Early modeling
studies of the Gaseous Electronics Conference (GEC) reference cell established numerical descriptions of the
bulk, sheath, and radial discharge
structure~\cite{govindan1995gec,boeuf1995gec,lymberopoulos1995gec,turner2013benchmark}.
More comprehensive equipment models, including the Hybrid Plasma
Equipment Model (HPEM), have further incorporated electron
kinetics, heavy species transport, electromagnetic fields, and surface
reactions to connect reactor conditions with plasma behavior and wafer-facing
fluxes~\cite{kushner2009hybrid,yang2010maxwell,zhang2000surface}.

Recent developments in massively parallel fluid solvers, GPU implementations,
and dedicated acceleration schemes have improved the efficiency of
multidimensional plasma
simulations~\cite{chen2021parallel,shin2022gpu,li2025fast}. Nevertheless, CCP
simulations remain computationally demanding. They require coupling the
electric field with charged particle transport, resolving disparate spatial
and temporal scales, and advancing the solution over many radio-frequency (RF)
cycles until a periodic steady state is reached. A separate simulation is
required for each combination of operating parameters. Changing the electrode
gap introduces an additional complication because it alters both the
computational domain and mesh. 
{For the dataset generated in this study, the average
computational cost was $5.32$ CPU h per converged simulation.
This per-case cost accumulates across the many parameter combinations
needed for process-recipe optimization and reactor design, making
extensive simulation-based exploration computationally expensive.}

Machine learning has been applied to interpret plasma diagnostic data and
forecast plasma behavior from experimental
measurements~\cite{mesbah2019machine,bonzanini2023foundations,katesharbeck2019predicting}, as well
as to model plasma kinetics, chemistry, and plasma--surface
interactions~\cite{anirudh2023review,trieschmann2023machine}. Beyond these applications,
data-driven surrogate models reduce repeated computations in plasma simulation
by learning the relation between operating conditions and plasma response.
Recent studies have used such models to predict plasma evolution and spatially
resolved coupled fields~\cite{gopakumar2024plasma,liu2024plasma_surrogate}.
Deep transfer operator learning has also been used to predict
sheath-region plasma parameters in inductively coupled
reactors~\cite{ahn2025sheath}.
Surrogate predictions have also been incorporated into process optimization and
digital twin frameworks to accelerate process
development~\cite{ko2023optimization,myung2026digital} and reduce
reliance on repeated physical experiments~\cite{osowiecki2024sustainability}.

Many existing plasma surrogate models focus on predicting responses to varying operating conditions within a fixed reactor geometry. A few studies have begun
to account for geometric variation, including transfer between predefined
plasma-processing domains~\cite{liu2024plasma_surrogate}, learned Poisson
solvers for complex reactor geometries~\cite{siffa2024poisson}, parametric
models for tokamak equilibria~\cite{jang2024grad}, and predictions
of plasma quantities near the wafer for different edge ring
configurations~\cite{ahn2025sheath}. These studies demonstrate the feasibility
of incorporating geometric variation into plasma surrogate models.

Surrogate modeling for variable geometries is challenging because changes in
geometry may alter the number and locations of mesh points. By contrast, many
surrogate architectures assume a common spatial discretization and fixed input
and output dimensions across samples. To enable surrogate modeling across
variable geometries, several geometry-representation strategies can be used:
mapping each physical domain to a prescribed
reference domain or learned latent coordinates~\cite{yin2024dimon,li2023geofno},
embedding all geometries in a fixed extended domain~\cite{liu2023dafno},
representing fields as continuous functions of spatial coordinates, or directly
processing each geometry dependent physical discretization as a variable size
point set~\cite{li2020gno,zeng2025pcno,han2026geometric,huang2026cost,wu2024transolver}.
{
Although these strategies enable surrogate modeling across different
geometries, their comparative training-data requirements remain unclear
for CCP discharges with varying electrode gaps. This comparison should
consider both coupled plasma fields and wafer-adjacent ion-density
profiles, because low whole-field errors do not necessarily imply
accurate predictions near the wafer. Quantifying these requirements
is important for controlling reference-data generation costs:
at the per-case simulation cost reported above, a few hundred additional
training cases incur additional costs on the order of $10^3$ CPU h.
}

The main contributions of this work are summarized as follows:
\begin{enumerate}
    \item We systematically compare six geometry-aware surrogate models for coupled plasma field prediction in a GEC reference cell with varying electrode gaps.
    \item We quantify their training data requirements for accurate predictions of full plasma fields and ion density profiles near the wafer, providing practical guidance for model selection.
    \item We introduce charge-separation weighting into the training loss to improve predictions near the wafer and reduce reference data requirements.
\end{enumerate}

\section{Physical Model and Dataset Generation}

\subsection{Variable-Gap GEC Configuration and Fluid Model}
\label{sec:reference_plasma_model}

\begin{figure}[H]
 \centering
 \includegraphics[width=0.85\linewidth]{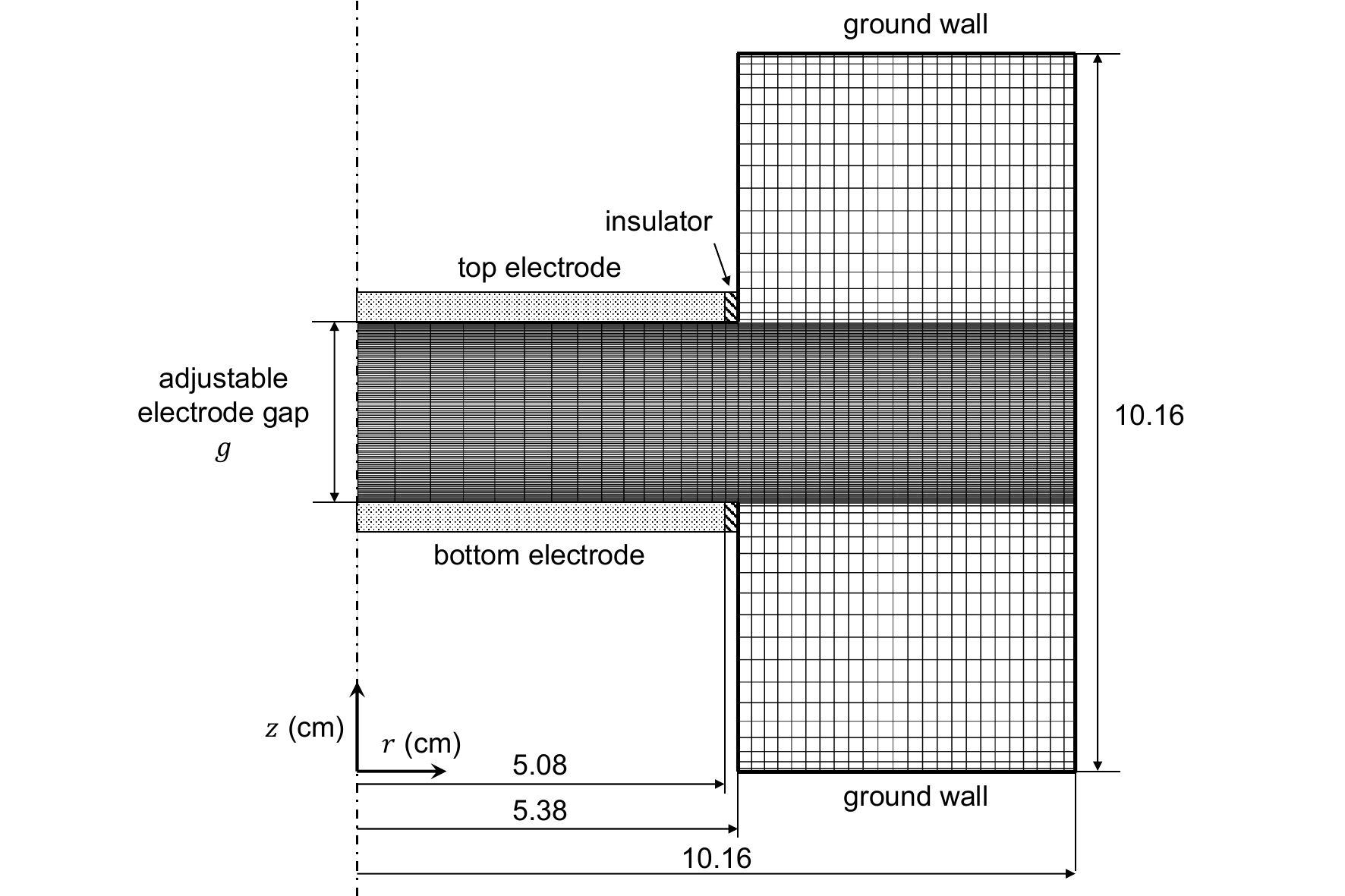}
 \caption{Two-dimensional axisymmetric computational domain and mesh of the
 variable-gap GEC configuration. The electrodes are separated from the
 surrounding grounded wall by thin annular insulators.}
 \label{fig:gec_configuration}
\end{figure}

The discharge is simulated in a two-dimensional axisymmetric
Gaseous Electronics Conference (GEC) reference cell
\cite{olthoff1995gec}, as illustrated in
Fig.~\ref{fig:gec_configuration}. The cylindrical chamber has a radius and
height of $10.16~\mathrm{cm}$ and contains two parallel circular electrodes
of radius $5.08~\mathrm{cm}$. Thin annular insulators separate the electrodes
from the surrounding chamber wall, whose inner edge is located at
$r=5.38~\mathrm{cm}$. All conductive chamber surfaces other than the two
electrodes are grounded. The axisymmetric condition is imposed at $r=0$.

For process optimization and reactor design, the following operating and
geometric parameters are of practical interest:
\begin{itemize}
 \item \textbf{Operating parameters:} Background argon pressure $p$, RF voltage amplitude
 $V_{\mathrm{RF}}$, and driving frequency $f_{\mathrm{RF}}$. At a fixed background gas temperature, $p$ determines the neutral argon density and affects collisions and
 transport. The electrodes are driven in a push--pull configuration by
 sinusoidal voltages with amplitude $V_{\mathrm{RF}}$, frequency
 $f_{\mathrm{RF}}$, and a phase difference of $\pi$,
 \begin{equation}
  \phi_b(t)=V_{\mathrm{RF}}\sin(2\pi f_{\mathrm{RF}}t),
  \qquad
  \phi_t(t)=V_{\mathrm{RF}}\sin(2\pi f_{\mathrm{RF}}t+\pi),
  \label{eq:push_pull_voltage}
 \end{equation}
 where $b$ and $t$ denote the bottom and top electrodes, respectively.

 \item \textbf{Geometric parameter:} The electrode separation $g$. The upper
 and lower electrodes move symmetrically about the chamber midplane at
 $z_c=5.08~\mathrm{cm}$, while the chamber dimensions and radial geometry
 remain unchanged. Their axial surface positions are
 \begin{equation}
  z_b(g)=z_c-\frac{g}{2},
  \qquad
  z_t(g)=z_c+\frac{g}{2}.
  \label{eq:electrode_positions}
 \end{equation}
\end{itemize}

For each combination of the operating and geometric parameters, the plasma
response is characterized by four time-dependent fields: electron density $n_e$, ion density $n_i$,
electron energy density $\mathcal{E}_e$, and electrostatic potential $\phi$.
Their dependence on position, time, operating conditions, and geometry is
written as
\begin{equation}
 \boldsymbol q(r,z,t;\,p,V_{\mathrm{RF}},f_{\mathrm{RF}},g)
 =\bigl(n_e,n_i,\mathcal{E}_e,\phi\bigr)^{\mathsf T}.
 \label{eq:plasma_field_dependence}
\end{equation}

These fields are obtained by solving a drift--diffusion fluid
model for electrons $e$ and singly charged argon ions $\mathrm{Ar}^{+}$.
The background gas consists of ground-state argon atoms $\mathrm{Ar}$ and is assumed to be stationary and spatially uniform at $T_g=300~\mathrm{K}$, with number density
$n_{\mathrm{Ar}}=p/(k_{\mathrm B}T_g)$.
Here $k_{\mathrm B}$ is the Boltzmann constant.
The electron and ion number densities satisfy
\begin{align}
 \frac{\partial n_e}{\partial t}
	+\nabla\cdot\boldsymbol{\Gamma}_e &= S_e,
	\qquad
	\boldsymbol{\Gamma}_e
	=-D_e\nabla n_e + \mu_e n_e\nabla\phi,
	\label{eq:continuity_electrons} \\
	\frac{\partial n_i}{\partial t}
	+\nabla\cdot\boldsymbol{\Gamma}_i &= S_i,
	\qquad
	\boldsymbol{\Gamma}_i
	=-D_i\nabla n_i - \mu_i n_i\nabla\phi,
	\label{eq:continuity_ion}
\end{align}
where $\boldsymbol{\Gamma}_e$ and $\boldsymbol{\Gamma}_i$ are the electron and ion particle fluxes,
and $D_s$ and $\mu_s$ denote the diffusion coefficient and mobility of species
$s$, respectively.
Since each electron-impact ionization event produces one electron--ion pair, 
the volumetric source terms are given by
\begin{equation}
 S_e=S_i
 =k_{\mathrm{ion}}(T_e)n_en_{\mathrm{Ar}}.
 \label{eq:ionization_source}
\end{equation}
Following the reduced electron-impact model of Ref.~\cite{boeuf1995gec}, the collision set retains ionization, excitation, and elastic scattering and is given by
\begin{align}
    e+\mathrm{Ar} &\rightarrow 2e+\mathrm{Ar}^{+},
    &\Delta\varepsilon_{\mathrm{ion}} &= 15.68~\mathrm{eV}, \nonumber\\
    e+\mathrm{Ar} &\rightarrow e+\mathrm{Ar}^{*},
    &\Delta\varepsilon_{\mathrm{ex}} &= 11.56~\mathrm{eV}, \nonumber\\
    e+\mathrm{Ar} &\rightarrow e+\mathrm{Ar},
    &\Delta\varepsilon_{\mathrm{el}} &= 3m_e k_{\mathrm B}(T_e-T_g)/m_{\mathrm{Ar}},
    \label{eq:argon_reactions}
\end{align}
where $m_e$ and $m_{\mathrm{Ar}}$ denote the electron and
argon-atom masses, respectively.
The excited state $\mathrm{Ar}^{*}$ is not transported as an independent species,
and the excitation process contributes only to the electron energy loss. The reduced
electron mobility $n_{\mathrm{Ar}}\mu_e$, reduced electron diffusion coefficient $n_{\mathrm{Ar}}D_e$, and the
ionization and excitation rate coefficients $k_{\mathrm{ion}}$
and $k_{\mathrm{ex}}$ are tabulated as functions of $T_e$.
These data were obtained from BOLSIG+ calculations~\cite{hagelaar2005bolsig} using
the Phelps electron--argon collision cross sections~\cite{yamabe1983argon} at
$300~\mathrm{K}$. The ion mobility and diffusion coefficient are parameterized by
the local reduced electric field $E/n_{\mathrm{Ar}}$, where
$E=|\nabla\phi|$ is the electric-field magnitude, using data reported by Ellis et
al.~\cite{ellis1976transport}.

The electron energy density $\mathcal{E}_e = \frac{3}{2}n_ek_{\mathrm B}T_e$ is governed by
\begin{equation}
 \frac{\partial \mathcal{E}_e}{\partial t}
 +\nabla\cdot\boldsymbol q_e
 -e\boldsymbol{\Gamma}_e\cdot\nabla\phi
 =-Q_{\mathrm{coll}},
 \qquad
 \boldsymbol q_e
 =-\frac{5}{2}k_{\mathrm B}D_en_e\nabla T_e
 +\frac{5}{2}k_{\mathrm B}T_e\boldsymbol{\Gamma}_e,
 \label{eq:electron_energy}
\end{equation}
where $\boldsymbol q_e$ is the electron energy flux and $e$
is the elementary charge.
The collisional energy-loss term is evaluated as
\begin{equation}
 Q_{\mathrm{coll}}
 =n_e\left[
 n_{\mathrm{Ar}}k_{\mathrm{ion}} (T_e) \Delta\varepsilon_{\mathrm{ion}}
 +n_{\mathrm{Ar}}k_{\mathrm{ex}} (T_e) \Delta\varepsilon_{\mathrm{ex}}
 +\frac{e}{m_e\mu_e}\Delta\varepsilon_{\mathrm{el}}
 \right].
 \label{eq:collisional_energy_loss}
\end{equation}
The electrostatic potential is coupled to the charged-species densities through the Poisson equation: 
\begin{equation}
 -\nabla\cdot\left(\varepsilon_0\nabla\phi\right)=e(n_i-n_e),
 \label{eq:poisson}
\end{equation}
where $\varepsilon_0$ is the vacuum permittivity.
The potential boundary conditions at the electrodes are given
by Eq.~\eqref{eq:push_pull_voltage}.
Absorbing drift--diffusion boundary conditions are imposed on all solid
surfaces~\cite{hagelaar2000boundary,sakiyama2007nonthermal}. With $\boldsymbol n$ denoting the outward unit normal of the plasma
domain, the electron and ion wall fluxes are
\begin{equation}
 \boldsymbol{\Gamma}_e\cdot\boldsymbol n
 =\frac{1}{4}n_e
 \sqrt{\frac{8k_{\mathrm B}T_e}{\pi m_e}}
 -\gamma\boldsymbol{\Gamma}_i\cdot\boldsymbol n, \qquad
 \boldsymbol{\Gamma}_i\cdot\boldsymbol n
 =\max\left(
 -\mu_i n_i\nabla\phi\cdot\boldsymbol n,\,0
 \right).
 \label{eq:wall_fluxes}
\end{equation}
The secondary-electron-emission coefficient is set to $\gamma=0$. The
electron energy flux at a solid surface is specified consistently as
\begin{equation}
 \boldsymbol q_e\cdot\boldsymbol n
 =\frac{5}{2}k_{\mathrm B}T_e
 \boldsymbol{\Gamma}_e\cdot\boldsymbol n.
 \label{eq:electron_energy_wall_flux}
\end{equation}

The governing equations are discretized using a cell-centered finite-volume
method on a nonuniform structured axisymmetric mesh. The mesh is locally
refined throughout the interelectrode discharge region, with additional
clustering toward the electrode surfaces and the radial electrode edge to
resolve the sheath and edge structures. Diffusive fluxes are evaluated using
central differencing, whereas the drift terms are discretized using a
first-order upwind scheme. Temporal derivatives in the charged-species
continuity and electron energy equations are discretized using the
first-order backward Euler scheme. At each time step, the electric potential
is first updated using a semi-implicit Poisson formulation
\cite{lin2012semiimplicit}, followed by
sequential implicit solutions of the electron continuity, ion continuity,
and electron energy equations. The resulting sparse linear systems are
solved using the SparseLU direct solver \cite{li2005overview}.
Each RF period is resolved using 200 uniform time steps. Each simulation is
advanced until a periodic steady state is reached, after which the plasma
quantities used for database construction are averaged over the final RF
period.
The two-dimensional GEC solver was previously benchmarked at a fixed electrode gap in AI4Plasma I~\cite{zou2026ai4plasma}.

\subsection{Gap-Dependent Plasma Characteristics}
\label{sec:gap_dependent_characteristics}

Experimental and numerical studies of CCP discharges have
extensively examined the effects of gas pressure, RF voltage amplitude,
and driving frequency on plasma density, electron heating, and sheath
characteristics~\cite{govindan1995gec,lymberopoulos1995gec,godyak1992highvoltage,zhu2007frequency,sharma2018control}.
In AI4Plasma I, we examined how plasma properties varied with these operating parameters at fixed electrode spacing~\cite{zou2026ai4plasma}.
Geometric studies have further shown that the effects of electrode spacing
on radial plasma distributions and electrode ion-current uniformity depend
on the operating conditions~\cite{boeuf1995gec}.
To identify the physical features that the surrogate models need to
capture, we examine the plasma response to electrode-gap variations from
two complementary perspectives:
\begin{itemize}
    \item \textbf{Discharge structure and mean plasma properties.}
    Axial profiles of ion density, electrostatic potential, and electron
    temperature, together with their volume-weighted averages within the
    interelectrode discharge, characterize the sheath structure and bulk
    plasma properties.
    \item \textbf{Radial ion-density distribution and uniformity.}
    Radial ion-density profiles are examined at the discharge midplane and
    near the wafer surface. Their area-weighted mean and coefficient of
    variation quantify the density level and radial nonuniformity,
    respectively.
\end{itemize}

\begin{figure}[H]
    \centering
    \includegraphics[width=\textwidth]{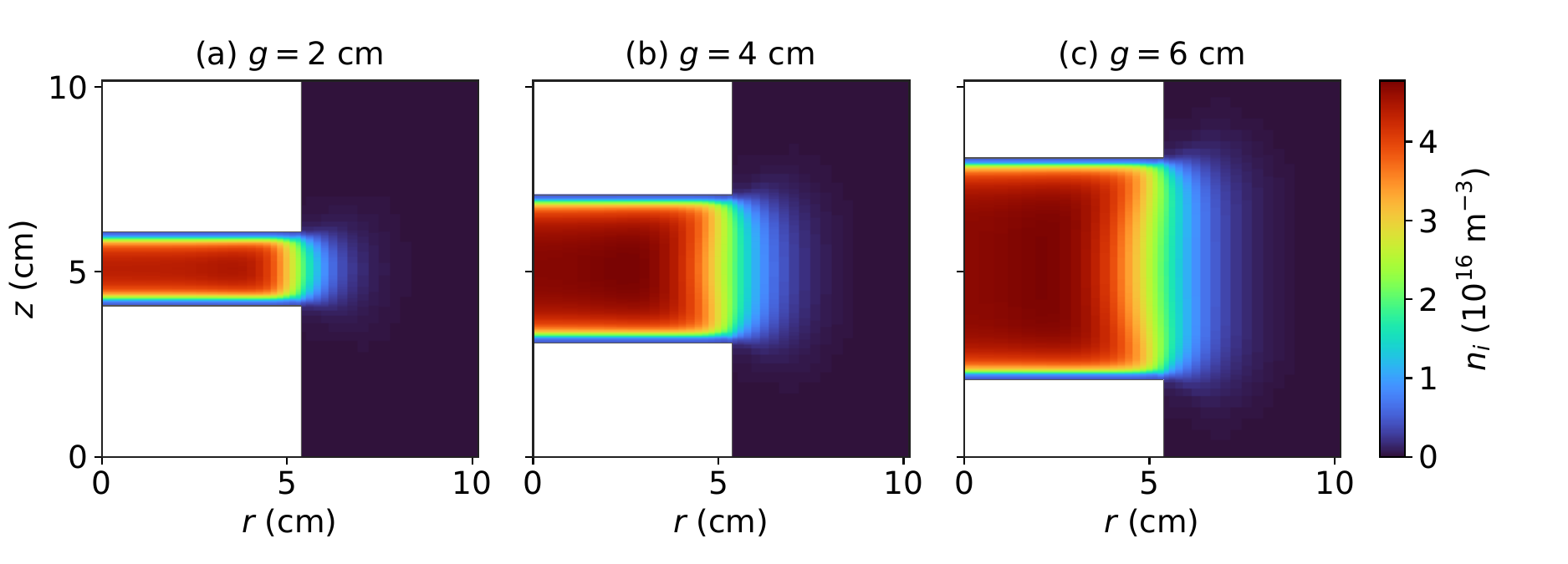}
    \caption{Ion-density distributions at electrode gaps of
    (a) 2, (b) 4, and (c) 6 cm, with $p=0.316~\mathrm{Torr}$,
    $V_{\mathrm{RF}}=200~\mathrm{V}$, and
    $f_{\mathrm{RF}}=40.68~\mathrm{MHz}$. All panels share the same color scale.}
    \label{fig:gap_ion_density_fields}
\end{figure}

Figure~\ref{fig:gap_ion_density_fields} shows that increasing
the electrode gap modifies both the axial extent and the radial distribution
of the plasma. At larger gaps, the ion density decreases more noticeably
from the discharge center toward the electrode edge, while the plasma
extends farther into the surrounding cavity.

\begin{figure}[H]
    \centering
    \includegraphics[width=\textwidth]{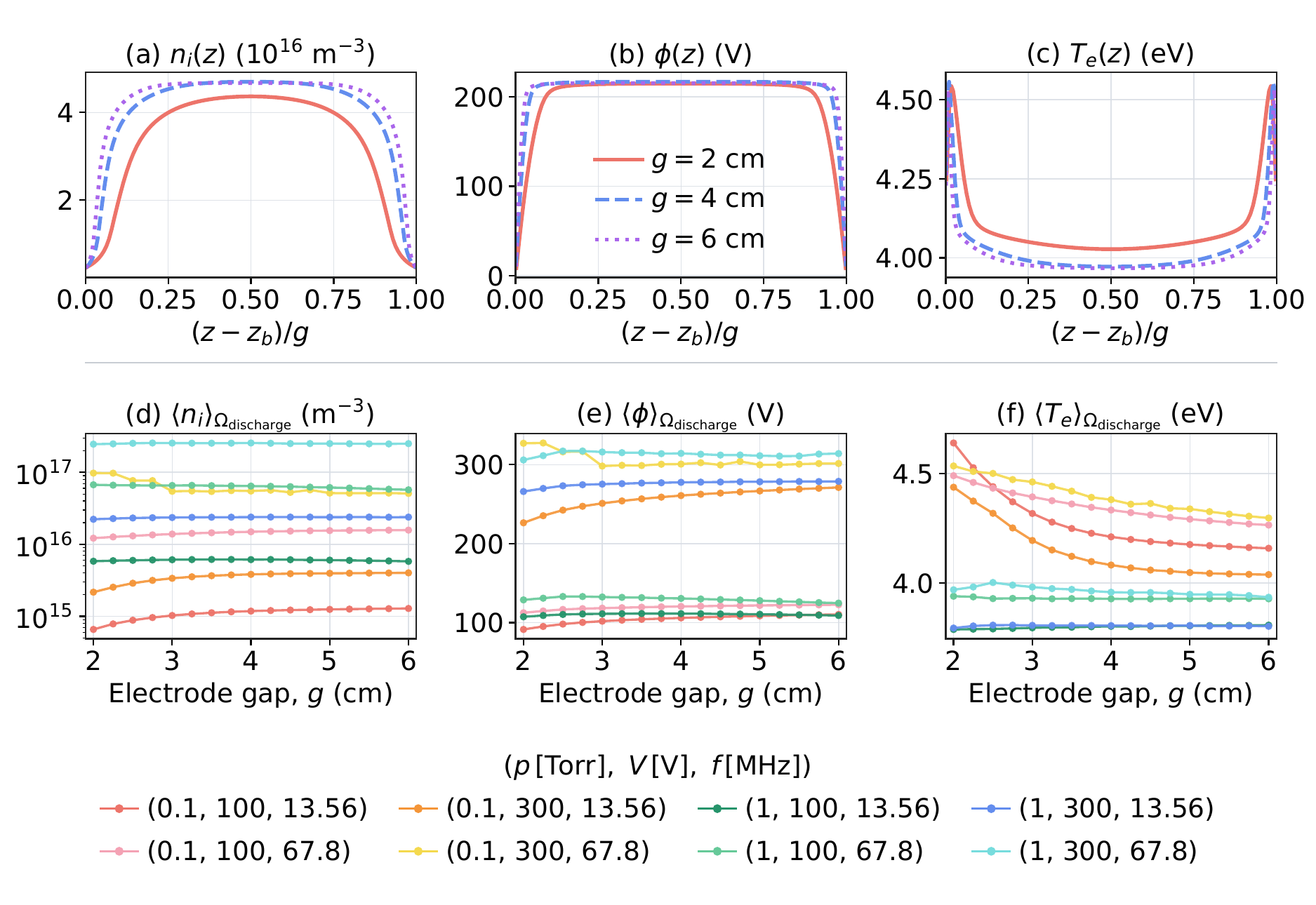}
    \caption{(a)--(c) Axial plasma profiles at
    $p=0.316~\mathrm{Torr}$, $V_{\mathrm{RF}}=200~\mathrm{V}$, and
    $f_{\mathrm{RF}}=40.68~\mathrm{MHz}$.
    (d)--(f) Discharge averages versus electrode gap at different operating
    conditions. Changes in the discharge-averaged ion density and potential
    are more pronounced at small gaps. Increasing the gap lowers the mean
    electron temperature, particularly at low pressure.}
    \label{fig:gap_physical_trends}
\end{figure}

Figure~\ref{fig:gap_physical_trends}(a)--(c) shows the axial profiles
near the symmetry axis, with the axial coordinate normalized as $(z-z_b)/g$ to facilitate comparison
among different electrode gaps. The profiles show that
increasing $g$ mainly reduces the relative extent of the near-electrode
sheath regions, while the bulk plasma properties remain comparatively
unchanged.
Figure~\ref{fig:gap_physical_trends}(d)--(f) presents the corresponding volume-weighted averages of these quantities over the interelectrode discharge region, defined as
\begin{equation}
	\Omega_{\mathrm{discharge}}(g)
	=
	\left\{
	(r,z)\in\Omega_g:
	0\le r\le R_e,\;
	z_b(g)\le z\le z_t(g)
	\right\}.
\end{equation}
The discharge-averaged $n_i$ is more sensitive to $g$ at
low pressure, particularly for small gaps. It increases with $g$ at the
lower voltage, whereas a pronounced decrease occurs at the higher voltage
and frequency. At high pressure, its gap dependence is comparatively
weak. The mean $\phi$ also shows more pronounced changes at small gaps,
with weaker changes at larger separations.
The mean $T_e$ decreases with increasing gap at low pressure
but changes only weakly at high pressure.

\begin{figure}[H]
    \centering
    \includegraphics[width=\textwidth]{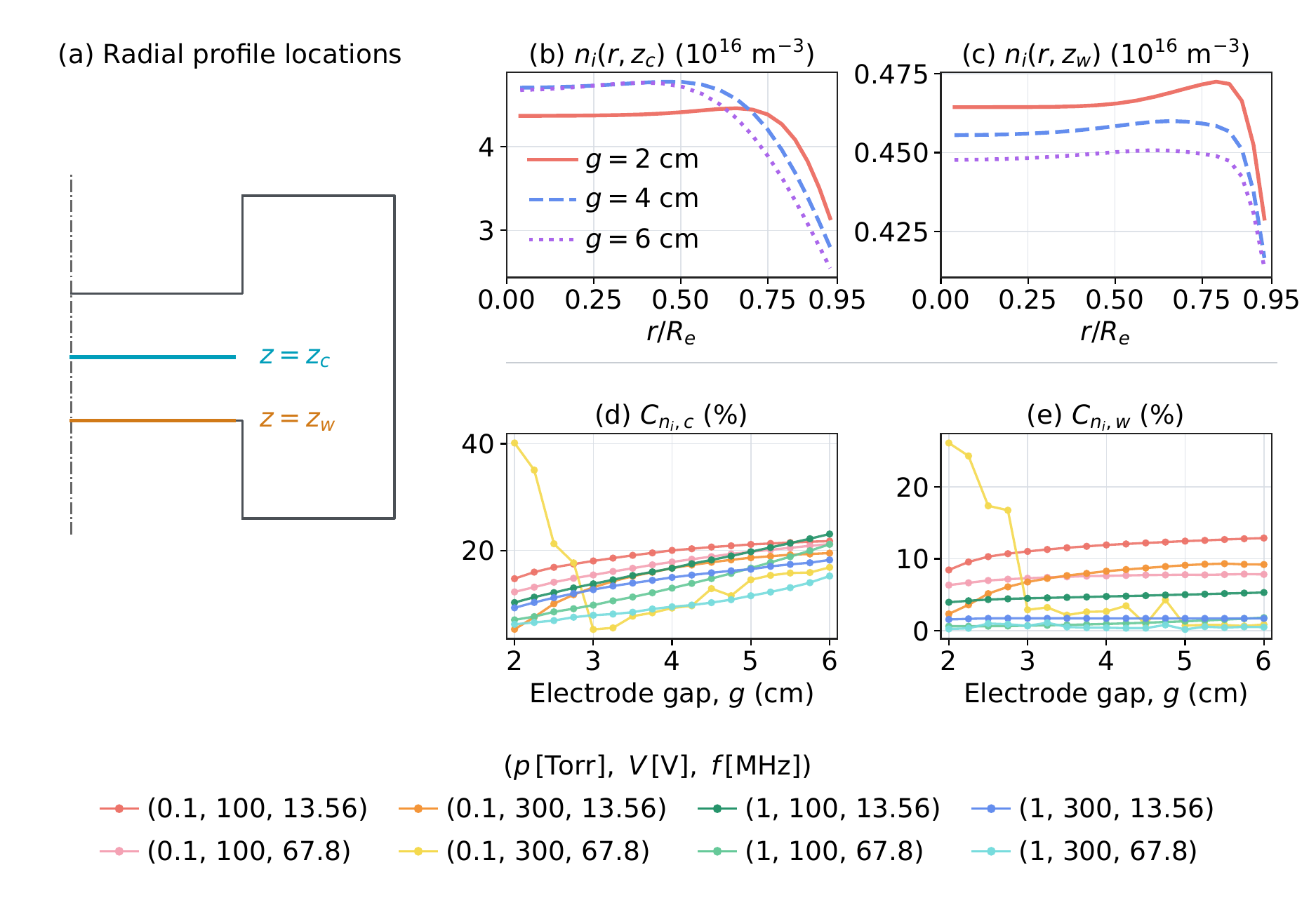}
    \caption{(a) Radial profile locations.
    (b), (c) Midplane and wafer-adjacent ion-density profiles at the same
    operating parameters as in Fig.~\ref{fig:gap_physical_trends}.
    (d), (e) Radial nonuniformity versus electrode gap at different operating
    conditions. Increasing the gap generally enhances midplane
    nonuniformity, while its effect near the wafer is weaker and more
    dependent on the operating conditions.}
    \label{fig:density_uniformity_characteristics}
\end{figure}

Figure~\ref{fig:density_uniformity_characteristics}(a)
indicates the two radial profile locations: the discharge midplane,
$z_c=[z_b(g)+z_t(g)]/2$, and the wafer-adjacent plasma layer, denoted by $z_w$. For an ion-density profile at either axial
location, the area-weighted radial mean and coefficient of variation, used here as a measure of radial nonuniformity, are defined as
\begingroup
\begin{equation}
 \left\langle n_i\right\rangle_r(z)
 =
 \frac{\displaystyle\int_0^{0.95R_e}n_i(r,z)\,2\pi r\,\mathrm{d}r}
 {\displaystyle\int_0^{0.95R_e}2\pi r\,\mathrm{d}r},
 \quad
 C_{n_i}(z)
 =
 \frac{\sqrt{\left\langle
 \left[n_i(r,z)-\left\langle n_i\right\rangle_r(z)\right]^2
 \right\rangle_r}}
 {\left\langle n_i\right\rangle_r(z)}\times100\%.
 \label{eq:radial_ni_descriptors}
\end{equation}
\endgroup
The outermost 5\% of the electrode radius is excluded from these scalar
statistics to reduce sensitivity to the electrode-edge singularity.

For the operating condition shown in
Fig.~\ref{fig:density_uniformity_characteristics}(b) and (c), increasing
the gap strengthens the center-to-edge density variation at the midplane.
Near the wafer, the density decreases, while the radial profile retains
a similar shape.
Figure~\ref{fig:density_uniformity_characteristics}(d)
and (e) shows that midplane nonuniformity generally increases with the
electrode gap, whereas wafer-adjacent nonuniformity varies more weakly
for most operating conditions. For the case with $p=0.1~\mathrm{Torr}$,
$V_{\mathrm{RF}}=300~\mathrm{V}$, and $f_{\mathrm{RF}}=67.8~\mathrm{MHz}$,
however, both locations exhibit a sharp reduction in nonuniformity over
the small-gap range.

\subsection{Dataset Generation}
\label{sec:dataset_generation}

Parameter combinations are selected within the ranges
summarized in Table~\ref{tab:dataset_parameter_ranges} using a maximin
Latin hypercube design~\cite{morris1995exploratory}. Sampling is performed
in the normalized space of electrode gap, logarithmic pressure, RF voltage
amplitude, and driving frequency to provide space-filling coverage across
the parameter range.

\begin{table}[H]
 \centering
 \caption{Operating-parameter ranges used for dataset generation.}
 \label{tab:dataset_parameter_ranges}
 \begin{tabular}{lccc}
  \toprule
  Parameter & Symbol & Interval & Unit \\
  \midrule
  Gas pressure & $p$ & [0.10, 1.00] & $\mathrm{Torr}$ \\
  RF voltage amplitude & $V_{\mathrm{RF}}$ & [100, 300] & $\mathrm{V}$ \\
  Driving frequency & $f_{\mathrm{RF}}$ & [13.56, 67.8] & $\mathrm{MHz}$ \\
  Electrode separation & $g$ & [2.0, 6.0] & $\mathrm{cm}$ \\
  \bottomrule
 \end{tabular}
\end{table}

Each parameter combination defines an independent plasma
simulation with the remaining physical and numerical settings fixed. The
cycle-averaged fields are obtained using the procedure described in
Section~\ref{sec:reference_plasma_model}. The database comprises 1000 training cases,
100 validation cases, and 100 test cases. These subsets are fixed before
model development and used consistently across all models.

The surrogate models take the four operating and geometric
parameters as inputs and predict the corresponding cycle-averaged plasma
fields,
\begin{equation}
    \overline{\boldsymbol q}(r,z;\,p,V_{\mathrm{RF}},f_{\mathrm{RF}},g)
    =\bigl(\overline n_e,\overline n_i,\overline{\mathcal E}_e,
    \overline\phi\bigr)^{\mathsf T}.
    \label{eq:surrogate_task}
\end{equation}
Hereafter, the temporal averaging overbars are omitted for brevity.
The reference fields are stored on the native solver grids
associated with each electrode gap. Their representation and preprocessing
for surrogate learning are introduced in the following section.

\section[Geometry Representations and Surrogate Architectures for Variable-Gap GEC Plasma]
{Geometry Representations and Surrogate Architectures for\\Variable-Gap GEC Plasma}

Developing a surrogate for a variable-gap GEC reactor is more challenging than
conventional parameter-to-field regression, because changing the electrode gap moves the internal boundaries of the computational domain.
A fixed spatial coordinate may lie in
the discharge bulk, in the sheath adjacent to an
electrode, or even outside the plasma region under different gap
configurations. Consequently, field patterns from different gap geometries
are not spatially aligned on a common grid. An appropriate representation of the variable geometry is therefore essential for learning the shared physical structures across different electrode gaps.

\begin{figure}[H]
    \centering
    \includegraphics[width=\textwidth]{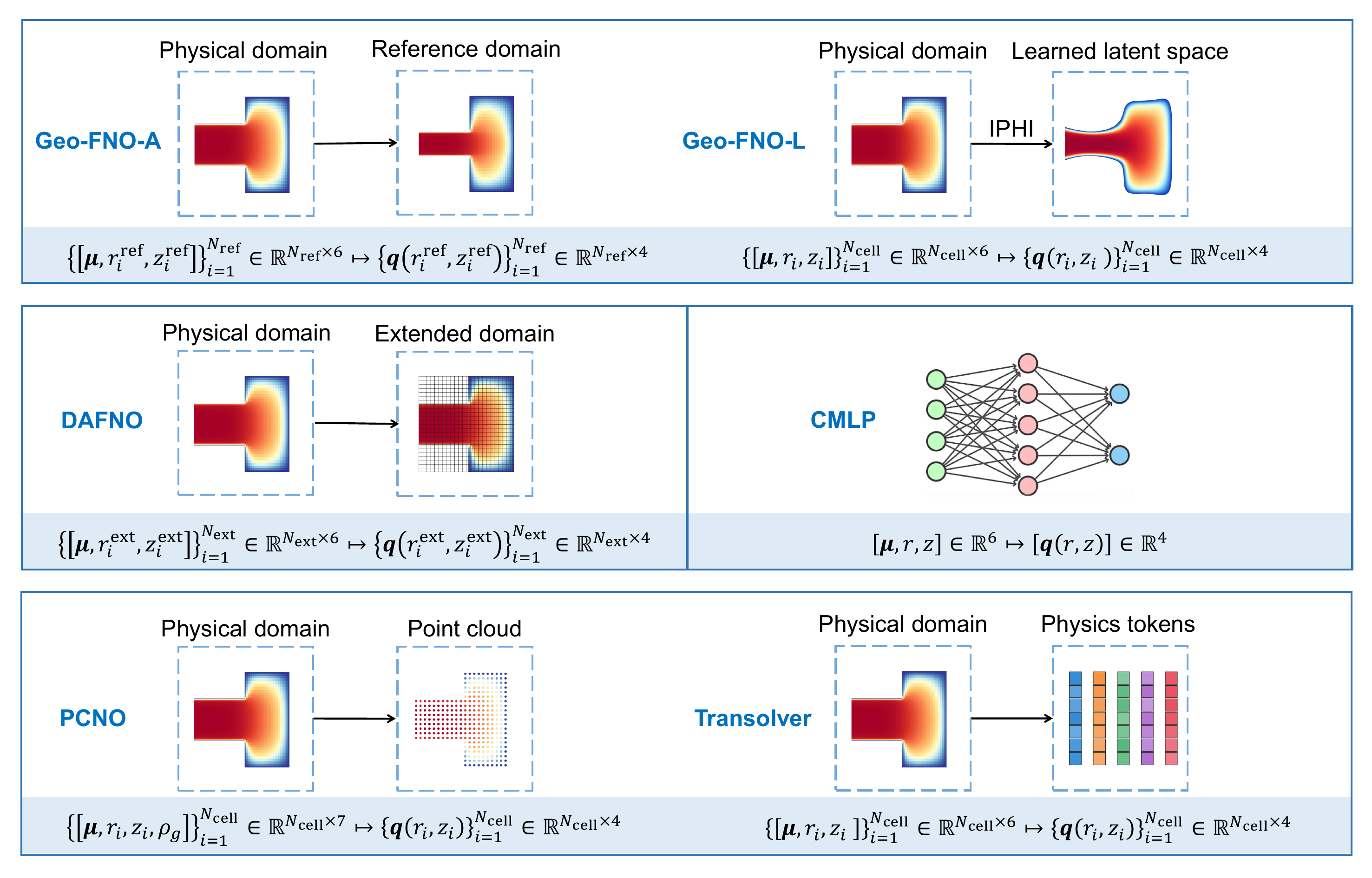}
    \caption{Overview of the geometry representations and corresponding surrogate architectures considered in this study.}
    \label{fig:geometry_representation_overview}
\end{figure}

This study compares six surrogate models spanning four geometry representation
strategies, as summarized in Fig.~\ref{fig:geometry_representation_overview}.
The model input parameters are denoted by
$\boldsymbol{\mu}=(g,\log_{10}p,V_{\mathrm{RF}},f_{\mathrm{RF}})$.
Geometry mapping aligns the physical domains through either a prescribed
transformation or a learned latent coordinate mapping. The extended-domain
representation places each geometry on a common rectangular grid and represents
its boundary explicitly. Continuous-coordinate representation models the fields
as continuous functions of the operating conditions and spatial coordinates.
Geometry-native discretizations instead operate directly on the variable-size
sets of active solver cells. The following subsections describe these
representations and their associated architectures, followed by the common
training and evaluation protocol.

\subsection{Geometry Mapping}

The Fourier neural operator (FNO)~\cite{li2021fno} performs spectral
convolution on a regular grid. After lifting and padding the input to a latent
representation, each residual Fourier block updates the feature field as
\begin{equation}
	\boldsymbol{h}^{(\ell+1)}
	=
	\boldsymbol{h}^{(\ell)}
	+
	\sigma\left(
	\mathcal{N}^{(\ell)}
	\left[
	\mathcal{F}^{-1}
	\left(
	\boldsymbol{R}^{(\ell)}
	\mathcal{F}\boldsymbol{h}^{(\ell)}
	\right)
	+
	\boldsymbol{W}^{(\ell)}\boldsymbol{h}^{(\ell)}
	\right]
	\right),
	\label{eq:fno_block}
\end{equation}
where $\mathcal{F}$ and $\mathcal{F}^{-1}$ denote the forward and inverse
Fourier transforms, $\boldsymbol{R}^{(\ell)}$ contains the learned weights
associated with the retained Fourier modes,
$\boldsymbol{W}^{(\ell)}$ denotes a pointwise channel transformation,
$\mathcal{N}^{(\ell)}$ is the normalization operation, and $\sigma$ is the
activation function.

The standard FNO cannot directly process the geometry-dependent solver meshes
because its Fourier operations are defined on a uniform structured grid. The
geometry-aware Fourier neural operator (Geo-FNO)~\cite{li2023geofno}
addresses this limitation by learning a mapping from the physical mesh to a
regular latent grid. In the learned-mapping configuration considered here, denoted Geo-FNO-L, the inverse-mapping network (IPHI) maps the physical coordinates to the latent coordinates according to
\begin{equation}
	\left(r_{\mathrm{lat}},z_{\mathrm{lat}}\right)
	=
	\mathcal{M}_{g}(r,z;\theta),
	\label{eq:learned_geo_mapping}
\end{equation}
where $\theta$ denotes the trainable parameters of the mapping network.
The learned transformation is incorporated into the leading and
trailing geometric spectral layers, while three intermediate Fourier layers
process the features on the regular latent grid.

To specialize Geo-FNO to the present variable-gap problem, we combine a
prescribed geometry mapping with a standard FNO and denote the resulting model
as Geo-FNO-A. The mapping aligns each physical domain $\Omega_g$ with a fixed
reference domain $\Omega_{\mathrm{ref}}$ corresponding to
$g_{\mathrm{ref}}=2.54~\mathrm{cm}$. It is defined by
$r_{\mathrm{ref}}=r$ and
\begin{equation}
	z_{\mathrm{ref}}=\mathcal{M}_g(z)=
	\begin{cases}
		z_{\min}+
		\dfrac{z-z_{\min}}{z_{\mathrm{b}}(g)-z_{\min}}
		\left[z_{\mathrm{b}}(g_{\mathrm{ref}})-z_{\min}\right],
		& z\le z_{\mathrm{b}}(g),\\[7pt]
		z_{\mathrm{b}}(g_{\mathrm{ref}})+
		\dfrac{z-z_{\mathrm{b}}(g)}{g}\,g_{\mathrm{ref}},
		& z_{\mathrm{b}}(g)<z\le z_{\mathrm{t}}(g),\\[7pt]
		z_{\mathrm{t}}(g_{\mathrm{ref}})+
		\dfrac{z-z_{\mathrm{t}}(g)}{z_{\max}-z_{\mathrm{t}}(g)}
		\left[z_{\max}-z_{\mathrm{t}}(g_{\mathrm{ref}})\right],
		& z>z_{\mathrm{t}}(g).
	\end{cases}
	\label{eq:geo_mapping}
\end{equation}
After applying the prescribed geometry mapping, the fields are interpolated onto a common structured grid over
$\Omega_{\mathrm{ref}}$. The aligned fields are processed on this grid using the FNO
blocks in Eq.~\eqref{eq:fno_block}, and the predictions are mapped back to the
original solver meshes for evaluation.

Geo-FNO-L learns the coordinate mapping, whereas Geo-FNO-A prescribes it from
the known geometry. The two models have comparable parameter counts and use
the same training protocol, allowing the two mapping strategies to be compared
under consistent settings.

\subsection{Extended-Domain Representation}

In the extended-domain representation, each gap-dependent physical domain
$\Omega_g$ is contained within the same fixed rectangular computational domain
\begin{equation}
 \mathcal{T}=[0,r_{\max}]\times[z_{\min},z_{\max}].
 \label{eq:embedded_box}
\end{equation}
The plasma fields on each original nonuniform solver mesh are interpolated
onto a uniform structured grid defined over $\mathcal{T}$. The uniform-grid resolution is selected according to the round-trip interpolation error between the original solver meshes and the candidate uniform grids. 
The lowest-cost grid with a mean round-trip error below $0.5\%$ is retained, giving $N_z\times N_r=250\times50$. Consequently, all
gap configurations share the same grid, while the plasma--solid boundary
varies within it.

Following this extended-domain preprocessing, the explicit domain-agnostic
Fourier neural operator (eDAFNO)~\cite{liu2023dafno} is adopted to incorporate
the physical-domain boundary directly into the operator layers. The geometry
is incorporated into every Fourier layer using a smoothed domain mask
constructed from the SDF,
\begin{equation}
 m_{\beta}(r,z)
 =\frac{1}{2}\left[1+\tanh\!\left(
 \beta\frac{d_{\mathrm{sdf}}(r,z)}{z_{\max}-z_{\min}}\right)\right],
 \label{eq:edafno_smooth_characteristic}
\end{equation}
where $\beta$ controls the transition width of the mask, and $d_{\mathrm{sdf}}(r,z)$ is the signed-distance function,
\begin{equation}
 d_{\mathrm{sdf}}(r,z)=
 \begin{cases}
  \phantom{-}d(r,z), & (r,z)\in\Omega_g,\\
  -d(r,z), & (r,z)\notin\Omega_g,
 \end{cases}
 \label{eq:embedded_sdf}
\end{equation}
where $d(r,z)$ is the minimum Euclidean distance to the plasma--solid
boundary. Let
$\boldsymbol{h}^{(\ell)}$ denote the latent field at layer $\ell$. The eDAFNO
layer is written as
\begin{align}
	\boldsymbol{h}^{(\ell+1)}(r,z)=\sigma\Bigg\{
	m_{\beta}(r,z)\Big[
	\mathcal{K}^{(\ell)}\!\left(
	m_{\beta}\boldsymbol{h}^{(\ell)}\right)(r,z)
	-\boldsymbol{h}^{(\ell)}(r,z)
	\mathcal{K}^{(\ell)}\!\left(m_{\beta}\right)(r,z)
	+\boldsymbol{W}^{(\ell)}
	\boldsymbol{h}^{(\ell)}(r,z)\Big]\Bigg\},
	\label{eq:edafno_layer}
\end{align}
where $\mathcal{K}^{(\ell)}$ is a learned translation-invariant convolution
operator, $\boldsymbol{W}^{(\ell)}$ is a learned pointwise transformation,
and $\sigma$ is the activation function. By incorporating the smoothed domain mask into each Fourier layer, eDAFNO restricts nonlocal interactions to the physical domain while retaining FFT-based evaluation. Compared with treating the mask or SDF only as an input channel, this formulation reduces information mixing across the plasma--solid boundary and reduces Gibbs oscillations associated with the boundary discontinuity.

\subsection{Coordinate-Based Continuous Representation}

The coordinate-based representation describes the plasma solution directly
as a continuous function of the operating conditions and physical coordinates.
The corresponding surrogate, denoted as C-MLP, is defined by the pointwise
 mapping
\begin{equation}
 \boldsymbol{X}_{\mathrm{C\text{-}MLP}}(r,z;\boldsymbol{\mu})
 =\left[\boldsymbol{\mu},r,z\right]
 \in\mathbb{R}^{6}
 \xrightarrow{\mathrm{MLP}}
 \boldsymbol{y}(r,z;\boldsymbol{\mu})
 =\left(y_e,y_i,y_{\mathcal E},y_{\phi}\right)^{\mathsf T}
 \in\mathbb{R}^{4},
 \label{eq:c_mlp_mapping}
\end{equation}
where $(r,z)\in\Omega_g$.

The variable geometry is represented implicitly by the electrode gap contained
in $\boldsymbol{\mu}$ together with the query coordinates $(r,z)$. Each
spatial location is then queried independently, so C-MLP requires neither
geometry mapping nor interpolation onto a common background grid. Unlike the
other models, however, it does not include an explicit spatial-interaction
operator.

\subsection{Geometry-Native Discretizations}

Geometry-native discretizations preserve the original spatial discretization
and apply the surrogate directly to the active cell centers, as in PCNO and
Transolver. For a case with $N_{\mathrm{cell}}$ active cells, the model processes the
variable-size point set
\begin{equation}
	\begin{aligned}
	\mathrm{PCNO}:\quad
	\left\{[\boldsymbol{\mu},r_i,z_i,\rho_g]\right\}_{i=1}^{N_{\mathrm{cell}}}
	&\in\mathbb{R}^{N_{\mathrm{cell}}\times7}
	\longmapsto
	\left\{\boldsymbol{y}_i\right\}_{i=1}^{N_{\mathrm{cell}}}
	\in\mathbb{R}^{N_{\mathrm{cell}}\times4},\\
	\mathrm{Transolver}:\quad
	\left\{[\boldsymbol{\mu},r_i,z_i]\right\}_{i=1}^{N_{\mathrm{cell}}}
	&\in\mathbb{R}^{N_{\mathrm{cell}}\times6}
	\longmapsto
	\left\{\boldsymbol{y}_i\right\}_{i=1}^{N_{\mathrm{cell}}}
	\in\mathbb{R}^{N_{\mathrm{cell}}\times4}.
	\end{aligned}
	\label{eq:native_point_set}
\end{equation}
Unlike C-MLP, which evaluates each point independently, both models process
the complete point set and exchange information among its spatial locations.
This formulation allows the direct processing of cases with different numbers
of input grid cells.

The point-cloud neural operator (PCNO)~\cite{zeng2025pcno} extends the FNO to
irregular point sets by evaluating the Fourier integral directly on the point
cloud and augmenting it with a local differential operator. Its input is given
in Eq.~\eqref{eq:native_point_set}, where
$\rho_g=L_{\mathrm{ref}}^3/\sum_jV_j$ is a dimensionless measure of the point-cloud
density, and 
$L_{\mathrm{ref}}=0.1016~\mathrm{m}$ is the coordinate-normalization length.
For the latent feature $\boldsymbol h^{(\ell)}$, a PCNO layer is written as
\begin{equation}
	\boldsymbol h_i^{(\ell+1)}
	=
	\sigma\left(
	\boldsymbol W^{(\ell)}\boldsymbol h_i^{(\ell)}
	+
	\mathcal K_{\mathrm F}^{(\ell)}[\boldsymbol h^{(\ell)}](\boldsymbol x_i)
	+
	\boldsymbol D^{(\ell)}\nabla_h\boldsymbol h_i^{(\ell)}
	\right),
	\label{eq:pcno_layer}
\end{equation}
where $\boldsymbol{x}_i=(r_i,z_i)$. The three terms inside the activation
represent pointwise feature transformation, global interaction through the
Fourier integral operator, and local interaction through the differential
operator, respectively. The Fourier integral is evaluated on the irregular
point set as
\begin{equation}
	\mathcal{K}_{\mathrm F}^{(\ell)}[\boldsymbol h^{(\ell)}](\boldsymbol x_i)
	\approx
	\sum_{j=1}^{N_{\mathrm{cell}}}
	\kappa^{(\ell)}(\boldsymbol x_i,\boldsymbol x_j)
	\boldsymbol h_j^{(\ell)}\,\omega_j,
	\quad
	\omega_j = V_j / \sum_{k=1}^{N_{\mathrm{cell}}}V_k,
	\quad
	V_j=2\pi r_j\Delta r_j\Delta z_j.
	\label{eq:pcno_fourier_integral}
\end{equation}
Here, $\kappa^{(\ell)}$ is the learned Fourier kernel, while $\omega_j$ accounts
for the unequal cell volumes in the numerical integral.

Transolver~\cite{wu2024transolver} adapts the Transformer architecture to
physical-field prediction through Physics-Attention. Instead of applying
self-attention directly among all grid cells, it first softly groups their
features into $N_s$ learned physics slices. The input matrix
$\boldsymbol X\in\mathbb{R}^{N_{\mathrm{cell}}\times6}$ defined in
Eq.~\eqref{eq:native_point_set} is first lifted to the latent point features
$\boldsymbol H^{(0)}=\boldsymbol X\boldsymbol W_{\mathrm{in}}+\boldsymbol b_{\mathrm{in}}$.
At layer $\ell$, 
physics tokens are obtained as
\begin{equation}
	\boldsymbol Z^{(\ell)}
	=
	\frac{
		{\boldsymbol S^{(\ell)}}^{\mathsf T}\boldsymbol F^{(\ell)}
	}{
		{\boldsymbol S^{(\ell)}}^{\mathsf T}\boldsymbol 1
	},
	\label{eq:transolver_tokens}
\end{equation}
Here, $\boldsymbol S^{(\ell)}$ gives the slice assignments,
$\boldsymbol F^{(\ell)}$ contains the projected point features, 
\begin{equation}
	\boldsymbol S^{(\ell)}
	 =\operatorname{softmax}\!\left(
	\boldsymbol H^{(\ell)}\boldsymbol W_s^{(\ell)}/\tau^{(\ell)}
	\right), \qquad
	 \boldsymbol F^{(\ell)}
	 =\boldsymbol H^{(\ell)}\boldsymbol W_f^{(\ell)}
	\label{eq:transolver_assignments}
\end{equation}
Here, the $\boldsymbol W$ matrices are learned projections, and
$\tau^{(\ell)}$ controls the sharpness of the slice assignments. 
Standard QKV self-attention
updates these tokens, which are then mapped back to the grid cells using the
same assignment weights. Thus, global information is exchanged through the
physics tokens rather than directly between all pairs of grid cells.

\subsection{Model Training}
\label{sec:model_training}

The electron energy density and charged-particle densities
span several orders of magnitude. To compress their dynamic ranges and
ensure positive predictions, all models predict a transformed field
$\boldsymbol y=(y_e,y_i,y_{\mathcal E},y_\phi)^{\mathsf T}$,
with the same component order as the physical field vector
$\boldsymbol q$ defined in Eq.~\eqref{eq:plasma_field_dependence}.
The physical fields are recovered through
\begin{equation}
	\begin{aligned}
		\mathcal{E}_e &= 10^{s_{\mathcal{E}}y_{\mathcal{E}}+m_{\mathcal{E}}}, &
		n_i &= 10^{s_i y_i+m_i},\\
		n_e &= 10^{s_e y_e+m_e}, &
		\phi &= s_{\phi}y_{\phi}+m_{\phi}.
	\end{aligned}
	\label{eq:four_field_transform}
\end{equation}
Here, $m$ and $s$ denote the means and standard deviations
of $\log_{10}\mathcal E_e$, $\log_{10}n_i$, and $\log_{10}n_e$ for the
three positive fields, and of $\phi$ for the potential. All transformation
statistics are computed using only the training data.

The architectural configurations are summarized in
Table~\ref{tab:model_architectures}. All models use the fixed dataset split
described in Section~\ref{sec:dataset_generation} and are trained with AdamW
using five paired random seeds. For each training case $k$, the loss is
the equal-weight mean of the axisymmetric volume-weighted relative squared
errors over the four physical fields,
\begin{equation}
\mathcal{L}^{(k)}
=\frac{1}{4}
\sum_{q\in\{\mathcal{E}_e,n_i,n_e,\phi\}}
\frac{\sum_j w_j^{(k)}
\left(\widehat q_j^{(k)}-q_j^{(k)}\right)^2}
     {\sum_j w_j^{(k)}\left(q_j^{(k)}\right)^2},
\label{eq:training_loss}
\end{equation}
where $q$ denotes an individual physical field,
$\widehat q$ denotes its surrogate prediction, and
$j$ indexes the active cells of case $k$. The weight
$w_j^{(k)}=2\pi r_j^{(k)}\Delta r_j^{(k)}\Delta z_j^{(k)}$ is the
corresponding axisymmetric cell-volume weight. The predicted physical
quantities $\widehat q$ are recovered from the transformed outputs before
computing the loss.

The training loss is obtained by averaging
$\mathcal L^{(k)}$ over the sampled cases. For C-MLP, the spatial sums are
estimated using uniformly sampled active cells during training, whereas
validation uses all active cells on the original solver grids. For all
models, the checkpoint score is the equal-weight mean of
$\mathcal L^{(k)}$ over the validation cases. This score is used for
learning-rate scheduling, early stopping, and checkpoint selection, while
the held-out test set is reserved for final evaluation.

\section{Results and Discussion}

The results presented in this section support Geo-FNO-A
as a practical starting point for surrogate modeling of variable-geometry
CCP discharges when a suitable geometry mapping is available, while stricter
accuracy requirements may favor Transolver. We first compare the six models
in terms of whole-field prediction accuracy and data efficiency, then
examine radial ion-density profiles and uniformity at the discharge midplane
and near the wafer. We further assess the effect of charge-separation
weighting in the training loss on near-wafer and whole-field accuracy.
These comparisons lead to the model-selection recommendations for specific
accuracy targets in Section~\ref{sec:practical_recommendations}.
Complementary sensitivity tests of geometry representation and surrogate
architecture are reported in Appendices~\ref{app:geometry_sensitivities}
and~\ref{app:architecture_sensitivities}, respectively.

\subsection{Overall Four-Field Accuracy and Data Efficiency}

We first compare the six surrogate models in terms of
four-field prediction accuracy. To evaluate models with different geometry
representations on a common basis, all predictions are converted back to
physical quantities and evaluated on the original solver grids. For each
physical field $q$ in test case $k$, the axisymmetric volume-weighted
relative $L_2$ error is defined as
\begin{equation}
\epsilon_q^{(k)}
=
\left[
\frac{\sum_j w_j^{(k)}
\left(\widehat q_j^{(k)}-q_j^{(k)}\right)^2}
     {\sum_j w_j^{(k)}\left(q_j^{(k)}\right)^2}
\right]^{1/2}\times100\%,
\label{eq:relative_l2}
\end{equation}
where $j$ indexes the active cells and
$w_j^{(k)}$ is the axisymmetric cell-volume weight defined in
Section~\ref{sec:model_training}. For each random seed, field-wise errors
are averaged over the 100 test cases, and the overall four-field error is
obtained by averaging the four field-wise errors with equal weight.
Results are reported as the mean and sample standard deviation across
five seeds.

\begin{table}[H]
    \centering
    \small
    \setlength{\tabcolsep}{4pt}
    \caption{Four-field relative $L_2$ errors on the original solver grids at $N=1000$.}
    \label{tab:overall_model_accuracy}
    \begin{tabular}{lccccc}
        \toprule
        Model & $\epsilon_{\mathcal E_e}$ (\%) &
        $\epsilon_{n_i}$ (\%) &
        $\epsilon_{n_e}$ (\%) &
        $\epsilon_{\phi}$ (\%) & Mean (\%) \\
        \midrule
        \multicolumn{6}{l}{\textit{Geometry mapping}} \\
        Geo-FNO-A    & $0.848\pm0.028$ & $0.864\pm0.027$ & $0.859\pm0.015$ & $1.046\pm0.029$ & $0.904\pm0.014$ \\
        Geo-FNO-L    & $0.961\pm0.024$ & $1.014\pm0.015$ & $0.981\pm0.020$ & $1.082\pm0.051$ & $1.010\pm0.022$ \\
        \addlinespace
        \multicolumn{6}{l}{\textit{Extended-domain representation}} \\
        eDAFNO       & $1.096\pm0.118$ & $1.166\pm0.101$ & $1.080\pm0.089$ & $1.940\pm0.069$ & $1.320\pm0.092$ \\
        \addlinespace
        \multicolumn{6}{l}{\textit{Continuous-coordinate representation}} \\
        C-MLP        & $1.531\pm0.091$ & $1.425\pm0.090$ & $1.547\pm0.105$ & $2.074\pm0.157$ & $1.644\pm0.110$ \\
        \addlinespace
        \multicolumn{6}{l}{\textit{Geometry-native discretizations}} \\
        Transolver   & $\mathbf{0.798\pm0.062}$ & $\mathbf{0.833\pm0.038}$ & $\mathbf{0.812\pm0.050}$ & $\mathbf{0.958\pm0.070}$ & $\mathbf{0.850\pm0.050}$ \\
        PCNO         & $0.836\pm0.046$ & $0.883\pm0.048$ & $0.848\pm0.050$ & $1.029\pm0.072$ & $0.899\pm0.052$ \\
        \bottomrule
    \end{tabular}
\end{table}

\textbf{Accuracy at sufficient data.} Table~\ref{tab:overall_model_accuracy} summarizes the
field-wise and overall errors for models trained on $N=1000$ cases.
Transolver achieves the lowest mean four-field error of $0.850\%$, with
PCNO and Geo-FNO-A close behind at approximately $0.90\%$. Geo-FNO-L is
slightly less accurate, while eDAFNO and C-MLP have higher errors,
particularly in the potential field. These results show that several
different geometry representations can attain approximately $1\%$
whole-field error when sufficient training data are available.

\begin{figure}[H]
    \centering
    \includegraphics[width=0.98\textwidth]{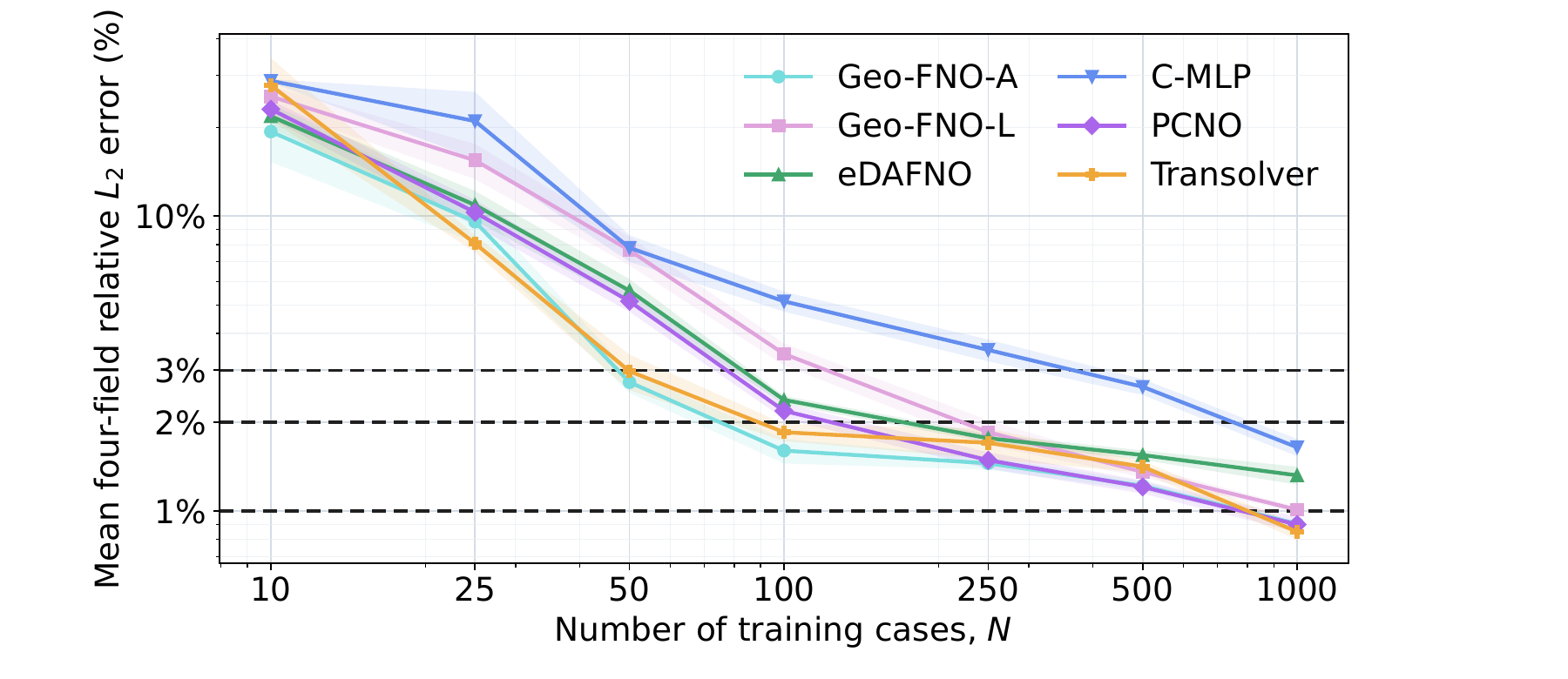}
    \caption{Data-efficiency comparison of the six surrogate models. Curves
    and shaded bands denote the mean and one sample standard deviation across
    five seeds. The horizontal dashed lines mark four-field mean-error levels
    of $1\%$, $2\%$, and $3\%$.}
    \label{fig:four_field_data_efficiency}
\end{figure}

\textbf{Data efficiency.} To further investigate model data efficiency, we use nested training subsets
with $N$ ranging from 10 to 1000 cases, while retaining the same validation and
test sets. Figure~\ref{fig:four_field_data_efficiency} reports the equal-weight
mean relative $L_2$ error over the four fields; the curves and bands denote the
mean and one sample standard deviation across five seeds. The horizontal
dashed lines mark reference error levels of $1\%$, $2\%$, and $3\%$.

Figure~\ref{fig:four_field_data_efficiency} shows that
Geo-FNO-A and Transolver have the lowest mean errors at $N=50$ and $100$,
whereas PCNO becomes competitive as the training set grows. To quantify
these differences in terms of reference-data requirements,
Table~\ref{tab:accuracy_target_training_sizes} summarizes the minimum
tested training-set sizes needed to reach the $3\%$, $2\%$, and $1\%$
error levels indicated by the dashed lines.
\begin{table}[H]
    \centering
    \caption{Minimum tested training-set sizes required to
    achieve $3\%$, $2\%$, and $1\%$ mean four-field relative $L_2$ error. Thresholds
    are determined from the mean errors across five random seeds. NR
    indicates that the target is not reached for $N\leq1000$. The smallest
    value in each column is highlighted in bold.}
    \label{tab:accuracy_target_training_sizes}
    \begingroup
    \setlength{\tabcolsep}{18pt}
    \begin{tabular}{lccc}
        \toprule
        Model & $N_{3\%}$ & $N_{2\%}$ & $N_{1\%}$ \\
        \midrule
        Geo-FNO-A & \textbf{50} & \textbf{100} & \textbf{1000} \\
        Geo-FNO-L & 250 & 250 & NR \\
        eDAFNO & 100 & 250 & NR \\
        C-MLP & 500 & 1000 & NR \\
        PCNO & 100 & 250 & \textbf{1000} \\
        Transolver & \textbf{50} & \textbf{100} & \textbf{1000} \\
        \bottomrule
    \end{tabular}
    \endgroup
\end{table}

At the $2\%$ target, Geo-FNO-A and Transolver require only
100 training cases, compared with 250 for Geo-FNO-L, eDAFNO, and PCNO, and
1000 for C-MLP. At the stricter $1\%$ target, only Geo-FNO-A, PCNO, and
Transolver reach the threshold, all at $N=1000$. Thus, model selection has
a substantial effect on reference-data requirements at the $2\%$ target,
whereas the tested training-set sizes do not distinguish the requirements
of the three models reaching $1\%$. With an average reference-simulation
cost of $5.32$ CPU h per case, reducing the training requirement from 1000
to 100 cases saves approximately $4.79\times10^3$ CPU h in data generation.

\begin{figure}[H]
    \centering
    \includegraphics[width=\textwidth]{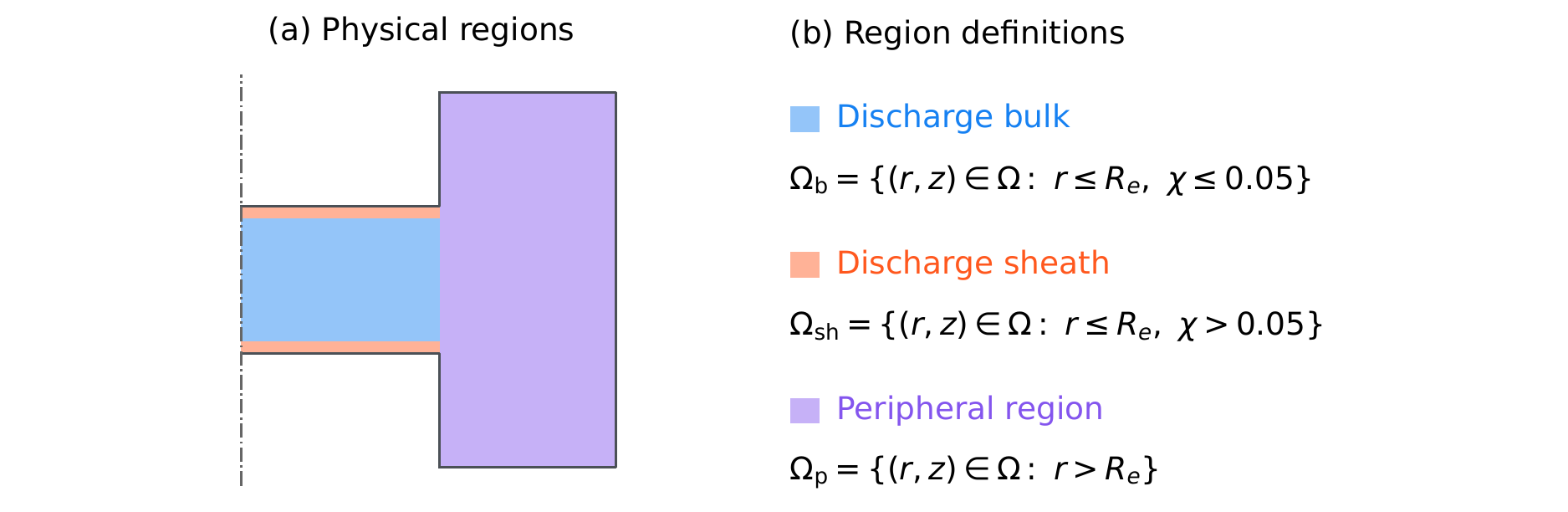}
    \caption{Physical regions used for error evaluation:
    (a) schematic partition of the GEC chamber and (b) region definitions.}
    \label{fig:physical_region_definition}
\end{figure}

\textbf{Accuracy across physical regions.} The whole-field errors above do not reveal how prediction
accuracy varies across the discharge. We therefore evaluate the errors
separately in the bulk, electrode-connected sheaths, and peripheral chamber
region, as illustrated in Fig.~\ref{fig:physical_region_definition}.
The bulk and sheath are distinguished using the normalized charge-separation
indicator of the reference fields,
\begin{equation}
 \chi=\frac{|n_i-n_e|}{n_i+n_e}.
 \label{eq:charge_separation_indicator}
\end{equation}
A threshold of $\chi=0.05$ identifies the sheath adjacent to
each electrode within the interelectrode discharge. The peripheral region
lies outside the electrode radius $R_e$. The region masks are determined
separately for each reference case and are shared by all models. Regional
errors are evaluated using Eq.~\eqref{eq:relative_l2}, with both sums
restricted to the corresponding region, and averaged with equal weight
over the four fields and 100 test cases within each seed.

\begin{figure}[H]
    \centering
    \includegraphics[width=0.98\textwidth]{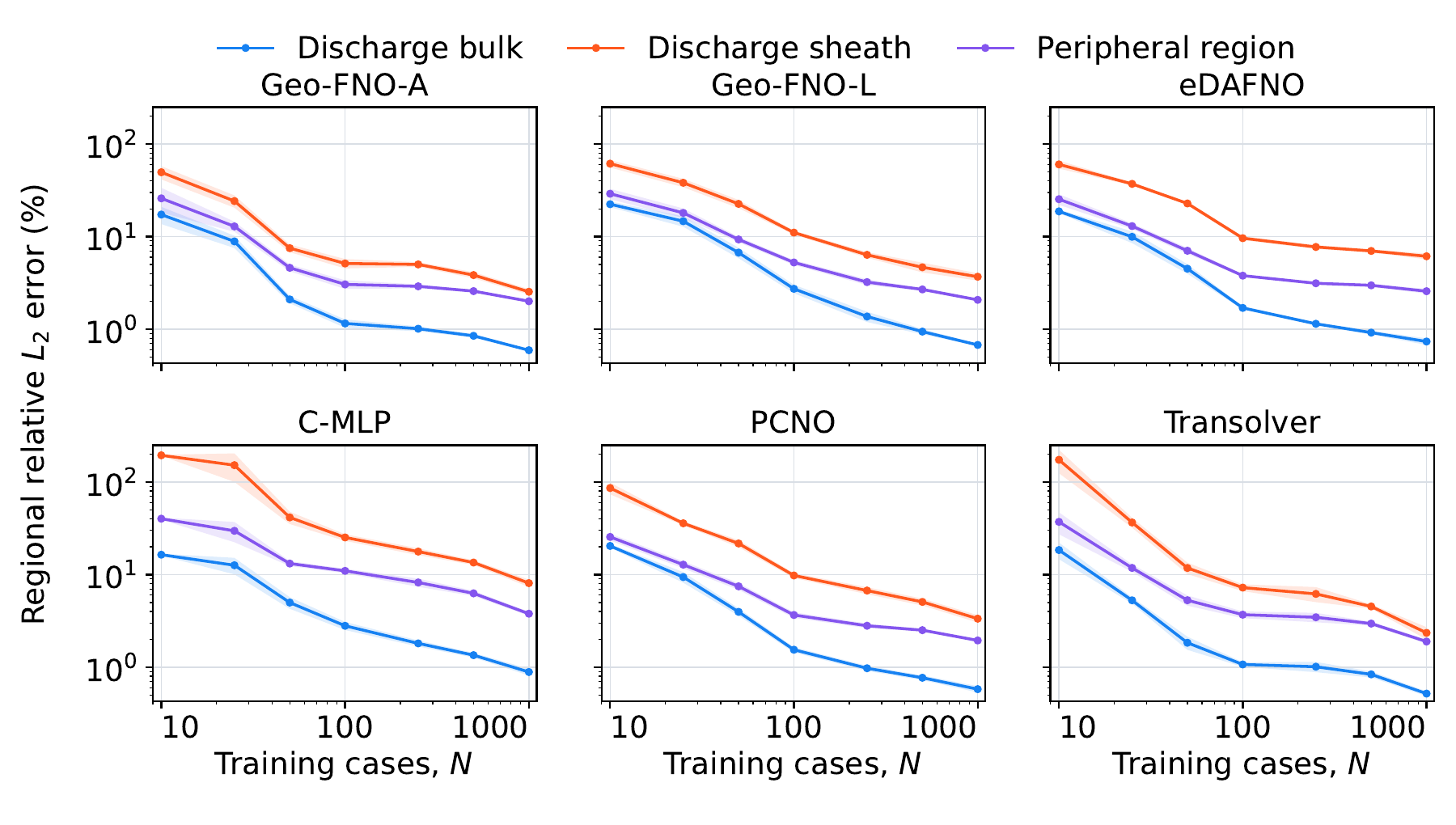}
    \caption{Regional four-field relative $L_2$ errors versus
    training-set size. Curves and shaded bands denote the mean and one
    sample standard deviation across five seeds, respectively, using the
    same 100 test cases at all training-set sizes. The sheath error remains
    higher than the bulk error as the training set grows.}
    \label{fig:physical_region_errors}
\end{figure}

Figure~\ref{fig:physical_region_errors} shows that increasing
the training-set size improves prediction accuracy in all three regions,
but a persistent accuracy gap remains between the bulk and the sheaths.
At $N=1000$, all six models achieve bulk errors below $1\%$, whereas the
sheath errors range from $2.35\%$ to $8.10\%$. The differences among models
are therefore more pronounced in the sheaths than in the bulk, with
Transolver and Geo-FNO-A giving the lowest sheath errors. Peripheral-region
errors generally lie between those of the bulk and the sheaths.
These regional differences show that low whole-field errors
do not imply equally accurate predictions near the electrodes. We therefore
examine the radial ion-density profiles near the wafer, including their mean
density and nonuniformity.

\subsection{Ion-Density Profile Accuracy}

To assess whether the models recover both the ion-density
level and radial uniformity near the wafer, we evaluate radial profiles at
different distances from the lower electrode. Three complementary metrics
measure the accuracy of the complete profile, its mean density, and its
nonuniformity. The profile $n_i(r;d)$ is extracted at distance
$d=z-z_b(g)$, from the electrode-adjacent sheath toward the discharge interior.

Consistent with Section~\ref{sec:gap_dependent_characteristics},
all radial metrics use the interval $0\leq r\leq0.95R_e$, excluding the
outermost $5\%$ to reduce sensitivity to the electrode-edge singularity.
The complete profile is evaluated using the area-weighted relative $L_2$ error,
\begin{align}
	\epsilon_{\mathrm{prof}}(d)
	&=
	\left[
	\frac{
		\displaystyle\int_0^{0.95R_e}
		\left[\widehat n_i(r;d)-n_i(r;d)\right]^2\;2\pi r\,\mathrm{d}r
	}{
		\displaystyle\int_0^{0.95R_e}
		n_i^2(r;d)\;2\pi r\,\mathrm{d}r
	}
	\right]^{1/2}
	\times100\%.
\end{align}
The radial mean $\langle n_i\rangle_r(d)$ and nonuniformity
$C_{n_i}(d)$ are evaluated using Eq.~\eqref{eq:radial_ni_descriptors}.
Their errors separately assess the predicted density level and radial
uniformity,
\begingroup
\begin{equation}
	\epsilon_{\mathrm{mean}}(d)=
	\frac{\left|\langle\widehat n_i\rangle_r(d)-\langle n_i\rangle_r(d)\right|}
	{\langle n_i\rangle_r(d)}\times100\%,
	\qquad
	\epsilon_C(d)=
	\left|\widehat C_{n_i}(d)-C_{n_i}(d)\right|.
\end{equation}
\endgroup
Here, $\epsilon_{\mathrm{mean}}$ is a relative error in
percent, whereas $\epsilon_C$ is an absolute difference in percentage points
(pp).

\begin{figure}[H]
    \centering
    \includegraphics[width=0.98\textwidth]{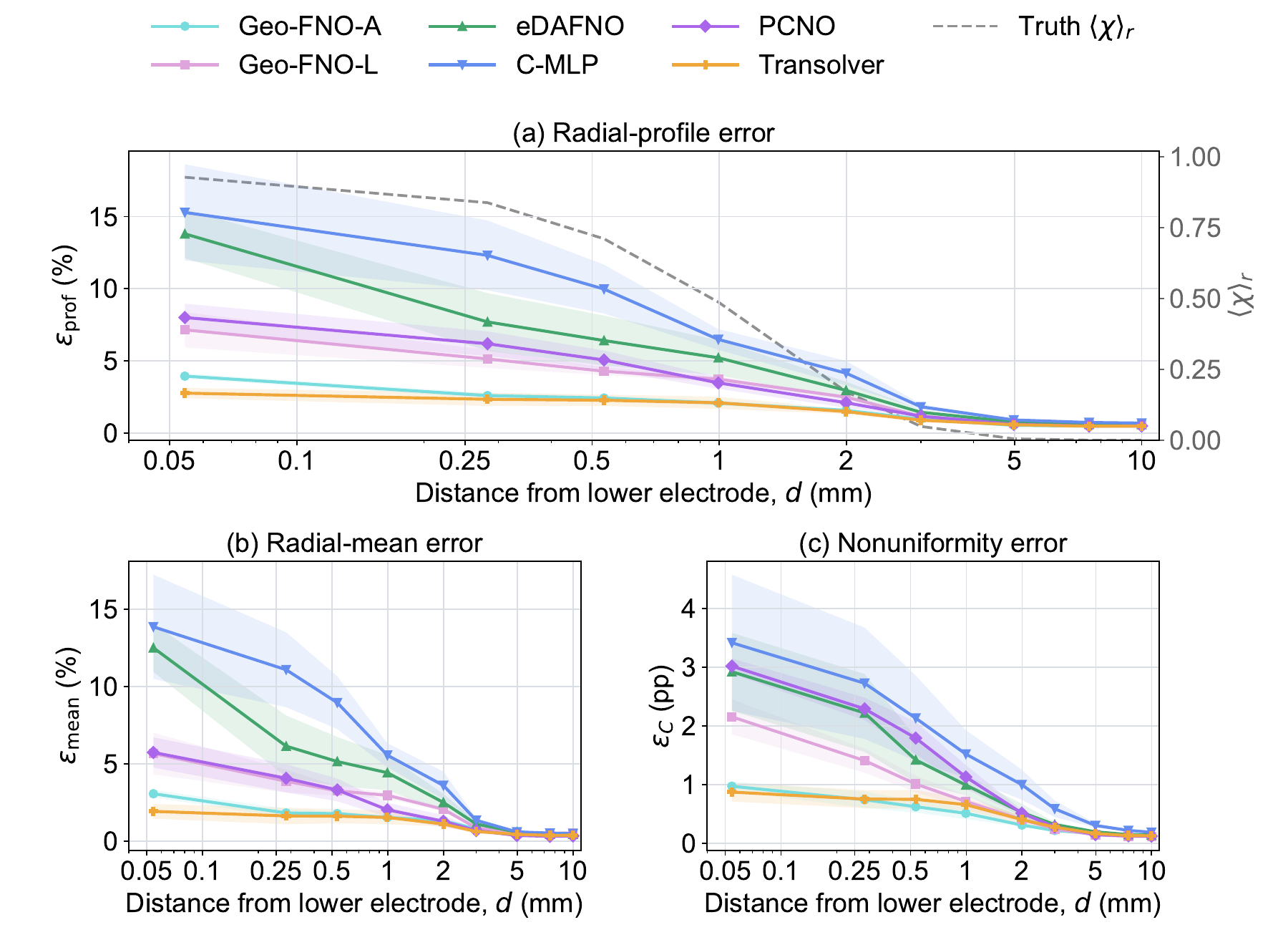}
    \caption{Ion-density prediction errors at $N=1000$ versus the distance
    $d=z-z_b(g)$ from the lower electrode: (a) radial-profile relative
    $L_2$ error, (b) radial-mean relative error, and (c) absolute
    nonuniformity error. Curves and shaded bands denote the mean and one sample
    standard deviation across five seeds, respectively. 
    The gray dashed line in panel (a) shows the radial-mean
    charge-separation indicator $\langle\chi\rangle_r$ of the reference
    solutions.}
    \label{fig:wafer_distance_ni_errors}
\end{figure}

\textbf{Ion-density profile accuracy.} Figure~\ref{fig:wafer_distance_ni_errors} shows that the
profile errors and differences among models decrease rapidly away from
the lower electrode. The largest errors occur in the electrode-adjacent
region, where the reference radial-mean charge-separation indicator
$\langle\chi\rangle_r$ is also high. Transolver and Geo-FNO-A give the
lowest near-electrode profile errors, approximately $2.8\%$ and $3.9\%$,
respectively, while the other models show larger deviations.
Panels (b) and (c) show that accurate prediction of the
mean density does not necessarily imply accurate radial uniformity. Near
the electrode, PCNO has a smaller mean-density error than eDAFNO and C-MLP,
but its nonuniformity error remains comparable to theirs. Geo-FNO-A and
Transolver maintain comparatively low errors in both quantities. The
differences in both metrics diminish toward the discharge interior.

Table~\ref{tab:ni_profile_accuracy} compares the prediction
errors at the discharge midplane and wafer-adjacent layer defined in
Fig.~\ref{fig:density_uniformity_characteristics}, using models trained
on $N=1000$ cases. All six models achieve midplane profile errors below
$0.8\%$, whereas their wafer-adjacent errors range from $2.76\%$ to
$15.29\%$. Transolver gives the lowest wafer-adjacent errors in all three
metrics, followed by Geo-FNO-A. Thus, comparable accuracy in the discharge
interior does not imply comparable recovery of the density level and
radial distribution near the wafer.

\begin{table}[H]
    \centering
    \caption{Mean ion-density prediction errors at the axial midplane and
    wafer-adjacent layer for $N=1000$. Values are averaged over 100 test cases
    and five seeds.}
    \label{tab:ni_profile_accuracy}
    \small
    \setlength{\tabcolsep}{6pt}
    \begin{tabular}{lcccccc}
        \toprule
        & \multicolumn{3}{c}{Axial midplane}
        & \multicolumn{3}{c}{Wafer-adjacent} \\
        \cmidrule(lr){2-4}\cmidrule(lr){5-7}
        Model
        & $\epsilon_{\mathrm{prof}}$ (\%) & $\epsilon_{\mathrm{mean}}$ (\%) & $\epsilon_C$ (pp)
        & $\epsilon_{\mathrm{prof}}$ (\%) & $\epsilon_{\mathrm{mean}}$ (\%) & $\epsilon_C$ (pp) \\
        \midrule
        Geo-FNO-A
        & $0.616$ & $0.518$ & $0.141$
        & $3.939$ & $3.065$ & $0.971$ \\
        Geo-FNO-L
        & $0.766$ & $0.647$ & $0.140$
        & $7.148$ & $5.668$ & $2.154$ \\
        eDAFNO
        & $0.672$ & $0.531$ & $0.148$
        & $13.804$ & $12.511$ & $2.924$ \\
        C-MLP
        & $0.729$ & $0.528$ & $0.189$
        & $15.288$ & $13.869$ & $3.417$ \\
        PCNO
        & $0.597$ & $\mathbf{0.465}$ & $\mathbf{0.119}$
        & $8.003$ & $5.739$ & $3.019$ \\
        Transolver
        & $\mathbf{0.581}$ & $0.473$ & $0.142$
        & $\mathbf{2.760}$ & $\mathbf{1.924}$ & $\mathbf{0.871}$ \\
        \bottomrule
    \end{tabular}
\end{table}

\begin{figure}[H]
    \centering
    \includegraphics[width=0.98\textwidth]{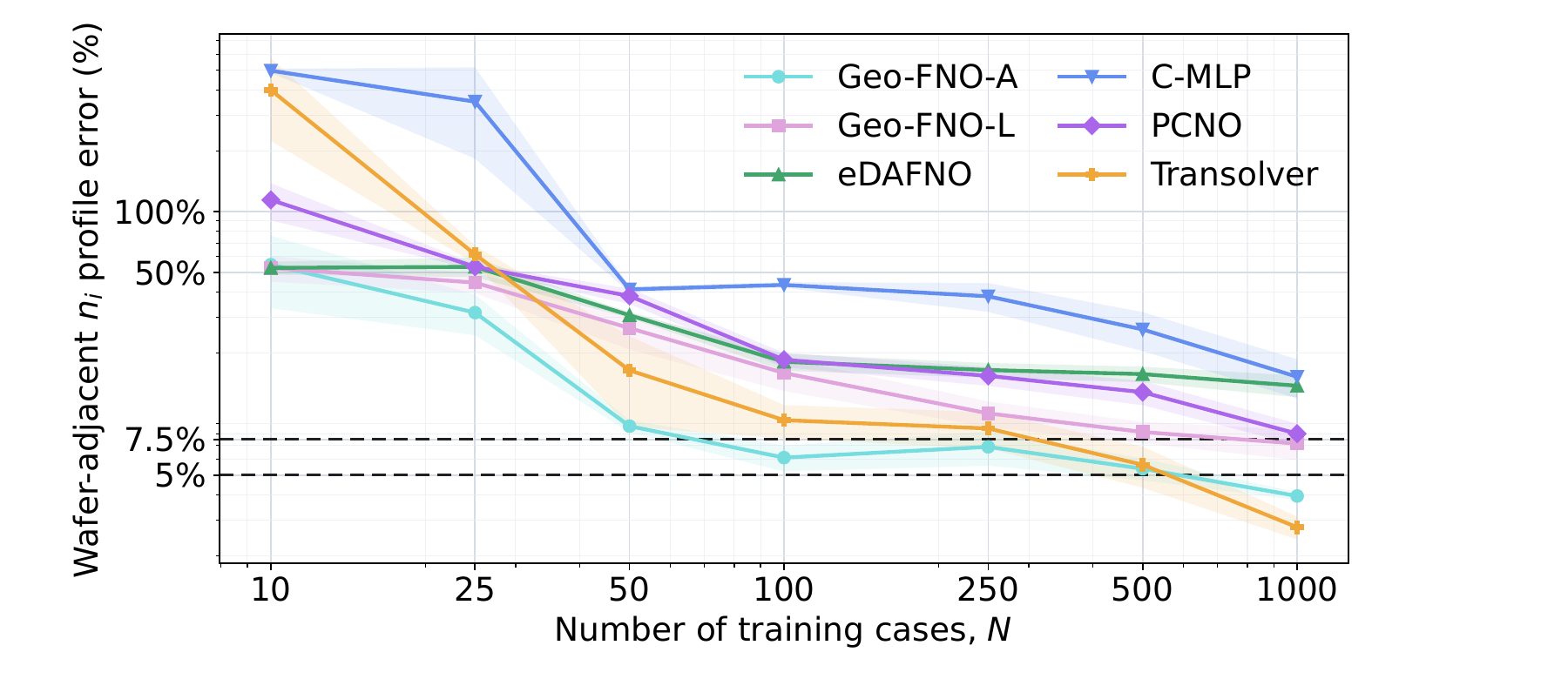}
    \caption{Wafer-adjacent ion-density radial-profile error versus training-set
    size. Curves and bands denote the mean and one sample standard deviation
    across five seeds. The horizontal dashed lines mark errors
    of $5\%$ and $7.5\%$.}
    \label{fig:wall_ni_accuracy}
\end{figure}

\textbf{Data efficiency.} Figure~\ref{fig:wall_ni_accuracy} examines the training-data
requirements for wafer-adjacent profile prediction. Errors decrease overall
as the training set grows, although the models differ substantially in
the number of reference cases needed to reach a prescribed accuracy.
Table~\ref{tab:wafer_accuracy_target_training_sizes} summarizes these
requirements at the $7.5\%$ and $5\%$ profile-error targets using the
minimum tested training-set sizes.
Geo-FNO-A reaches the $7.5\%$ target with 100 training cases,
	compared with 500 for Transolver and 1000 for Geo-FNO-L.
	Only Geo-FNO-A and Transolver reach the
	$5\%$ target within the tested range, both at $N=1000$.
	For comparison, these two models reach the whole-field $2\%$ target
	with only 100 training cases. Thus, the accuracy required near the wafer
	can impose a substantially greater reference-data requirement than
	the whole-field criterion.

\begin{table}[H]
    \centering
    \begingroup
    \caption{Minimum tested training-set sizes required to achieve
    $7.5\%$ and $5\%$ wafer-adjacent ion-density radial-profile relative $L_2$
    error. Thresholds are determined from the mean errors across five
    random seeds. NR indicates that the target is not reached for
    $N\leq1000$. The smallest value in each column is highlighted in bold.}
    \label{tab:wafer_accuracy_target_training_sizes}
    \setlength{\tabcolsep}{18pt}
    \begin{tabular}{lcc}
        \toprule
        Model & $N_{7.5\%}$ & $N_{5\%}$ \\
        \midrule
        Geo-FNO-A & \textbf{100} & \textbf{1000} \\
        Geo-FNO-L & 1000 & NR \\
        eDAFNO & NR & NR \\
        C-MLP & NR & NR \\
        PCNO & NR & NR \\
        Transolver & 500 & \textbf{1000} \\
        \bottomrule
    \end{tabular}
    \endgroup
\end{table}

\subsection{Charge-Separation-Weighted Loss}
\label{sec:charge_separation_regularization}
{ Errors in the thin wafer-adjacent region can be underemphasized
by a volume-weighted whole-field loss.}
To improve prediction accuracy near the wafer, we introduce
a charge-separation-weighted training loss that assigns greater importance
to errors in regions with appreciable charge separation. The modified loss
is obtained from Eq.~\eqref{eq:training_loss} by replacing its cell-volume
weights with
\begingroup
\begin{equation}
    \widetilde w_j^{(k)}
    =w_j^{(k)}\left(1+\lambda\,\chi_j^{(k)}\right).
    \label{eq:charge_separation_weights}
\end{equation}
\endgroup
Here, $\chi_j^{(k)}$ is the charge-separation indicator
defined in Eq.~\eqref{eq:charge_separation_indicator}, evaluated using the
reference training fields. {The parameter $\lambda\geq0$ controls the strength of the
additional weighting.} The modified weights are applied to both the
numerator and denominator of each field-wise relative squared error in
the training loss, and $\lambda=0$ recovers the original loss. {This weighting increases the relative contribution of errors in
charge-separated regions to the training objective, encouraging
more accurate predictions near the wafer.}

Geo-FNO-A, C-MLP, PCNO, and Transolver are trained with
either the original loss ($\lambda=0$) or the modified loss ($\lambda=4$)
at $N=100$ and $N=1000$, using five paired random seeds. The dataset splits,
model architectures, and validation criterion are retained from the
corresponding baselines. Prediction accuracy is evaluated on the same
100 test cases using the whole-field and wafer-adjacent profile metrics
defined above.

\begin{table}[H]
    \centering
    \begingroup
    \caption{Effect of the charge-separation-weighted training loss on prediction
    accuracy at $N=100$ and $N=1000$. Each entry reports the error for
    $\lambda=0\rightarrow\lambda=4$, averaged over 100 test cases and five
    paired random seeds.}
    \label{tab:charge_separation_weighting}
    \small
    \setlength{\tabcolsep}{9pt}
    \begin{tabular}{lcccc}
        \toprule
        & \multicolumn{2}{c}{$N=100$}
        & \multicolumn{2}{c}{$N=1000$} \\
        \cmidrule(lr){2-3}\cmidrule(lr){4-5}
        Model & \shortstack{Four-field\\error (\%)}
        & \shortstack{Wafer-adjacent $n_i$\\profile error (\%)}
        & \shortstack{Four-field\\error (\%)}
        & \shortstack{Wafer-adjacent $n_i$\\profile error (\%)} \\
        \midrule
        Geo-FNO-A & $1.601\rightarrow1.575$ & $6.094\rightarrow4.152$
                  & $0.904\rightarrow0.882$ & $3.939\rightarrow2.833$ \\
        C-MLP & $5.138\rightarrow5.353$ & $43.469\rightarrow38.920$
              & $1.644\rightarrow1.737$ & $15.288\rightarrow8.598$ \\
        PCNO & $2.185\rightarrow2.196$ & $18.551\rightarrow12.172$
             & $0.899\rightarrow0.867$ & $8.003\rightarrow4.394$ \\
        Transolver & $1.849\rightarrow1.785$ & $9.335\rightarrow5.901$
                   & $0.850\rightarrow0.858$ & $2.760\rightarrow2.094$ \\
        \bottomrule
    \end{tabular}
    \endgroup
\end{table}

Table~\ref{tab:charge_separation_weighting} shows that training
with the charge-separation weighted loss effectively reduces the mean wafer-adjacent profile error for
all four models at both training-set sizes. The relative reductions
range from $10.5\%$ to $36.8\%$ at $N=100$ and from $24.1\%$ to $45.1\%$
at $N=1000$. In particular, Geo-FNO-A reaches a profile error of $4.152\%$
with 100 training cases, meeting the $5\%$ target that its unweighted
baseline reaches only at $N=1000$ among the tested training-set sizes.
Changes in whole-field accuracy are comparatively small, although C-MLP
exhibits an increase from $1.644\%$ to $1.737\%$ at $N=1000$.
These results demonstrate that modifying the spatial
weighting in the training loss can improve near-wafer profile accuracy,
with a model-dependent trade-off in whole-field accuracy.
Despite the improvements in wafer-adjacent profile accuracy, the modified
loss preserves the model rankings in both whole-field and wafer-adjacent
profile accuracy at both training-set sizes.

\subsection{Practical Recommendations}
\label{sec:practical_recommendations}

With each reference simulation requiring an average of $5.32$ CPU h, model selection should account for the training data needed to achieve the desired accuracy. Table~\ref{tab:practical_model_characteristics} summarizes
the recommended models for different accuracy targets. Geo-FNO-A offers
a practical starting point for surrogate modeling of the variable-geometry
CCP discharges studied here, while stricter accuracy requirements may
favor Transolver. Two further points should be considered:
\begingroup
\begin{itemize}
    \item Incorporating charge-separation weighting into the training loss
    improves near-wafer profile accuracy. With this loss, Geo-FNO-A reaches
    the $5\%$ profile-error target with 100 training cases, compared with
    1000 for the original loss among the tested training-set sizes
    (Table~\ref{tab:charge_separation_weighting}).
    \item Geo-FNO-A requires an explicit geometry mapping, which may be
    difficult to construct for more complex chambers. Transolver may
    therefore be better suited to such geometries, as it operates directly
    on native point sets without requiring a prescribed mapping.
\end{itemize}
\endgroup

\begin{table}[H]
    \centering
    \begingroup
        \caption{Recommended surrogate models for different
    accuracy targets in the variable-gap GEC benchmark.
    Recommendations use the original training loss.}
    \label{tab:practical_model_characteristics}
    \small
    \setlength{\tabcolsep}{5pt}
    \renewcommand{\arraystretch}{1.15}
    \begin{tabular}{>{\raggedright\arraybackslash}p{0.28\textwidth}
                    >{\raggedright\arraybackslash}p{0.18\textwidth}
                    >{\raggedright\arraybackslash}p{0.47\textwidth}}
        \toprule
        Accuracy target & Recommended model & Supporting evidence \\
        \midrule
        Four-field mean error $\leq 2\%$
        & Geo-FNO-A
        & Reaches the target with $N=100$, as does Transolver, but yields
          a lower mean error ($1.601\%$ versus $1.849\%$). \\
        \addlinespace
        Four-field mean error $\leq 1\%$
        & Transolver
        & Reaches the target with $N=1000$, as do Geo-FNO-A and PCNO,
          and yields the lowest mean error ($0.850\%$). \\
        \addlinespace
        Wafer-adjacent profile error $\leq 7.5\%$
        & Geo-FNO-A
        & Reaches the target with $N=100$, compared with $N=500$
          for Transolver. \\
        \addlinespace
        Wafer-adjacent profile error $\leq 5\%$
        & Transolver
        & Reaches the target with $N=1000$, as does Geo-FNO-A, but yields
          a lower profile error ($2.760\%$ versus $3.939\%$). \\
        \bottomrule
    \end{tabular}
    \endgroup
\end{table}

\section{Conclusions}

This study compared six geometry-aware surrogate models for coupled plasma field prediction in a variable-gap axisymmetric GEC reference cell, evaluating accuracy and data requirements for both full plasma fields and radial ion density profiles. Geo-FNO-A and Transolver reach a mean four-field error of $2\%$ with 100 training cases, whereas the other models require 250--1000 cases. Geo-FNO-A performs most strongly when training data are limited, while Transolver achieves the highest accuracy with sufficient data. Differences become more pronounced near the wafer: with 1000 training cases, profile errors remain below $4\%$ for Transolver and Geo-FNO-A but exceed $13\%$ for eDAFNO and C-MLP, despite their comparable four-field accuracy. Thus, comparable accuracy in the full fields does not necessarily imply comparable accuracy near the wafer. Charge-separation weighting in the training loss improves ion density predictions near the wafer for all four models tested. For Geo-FNO-A, it reduces the minimum tested training requirement for a $5\%$ profile error from 1000 to 100 cases, saving approximately 4790 CPU h in reference simulations.

These results support Geo-FNO-A as a practical starting point when a suitable geometry mapping is available, and Transolver when sufficient training data are available and prediction accuracy is the priority. The present findings apply to cycle-averaged argon discharges with electrode gap as the geometric variable. Further work should examine broader reactor geometries, plasma chemistries, and operating regimes.

\section*{Acknowledgements}

This work was supported in part by Huawei Technologies Co., Ltd. Computational
resources and technical support were provided by the Theory Lab, Huawei Leibniz
Research Center. DZH acknowledges funding support of the National Key R\&D Program of China (No. 2025YFA1018700) and the NSFC (No. 62595771).

\section*{Competing interests}

Authors affiliated with Huawei Technologies Co., Ltd. include employees and
interns of the company. The authors declare no other competing financial
interests or personal relationships that could have influenced the work
reported in this paper.

\section*{Data availability}

The data that support the findings of this study are available from the
corresponding authors upon reasonable request.

\appendix
\section{Architectural Configurations of the Surrogate Models}
\label{app:model_architectures}
\setcounter{table}{0}
\renewcommand{\thetable}{A\arabic{table}}

\begin{table}[H]
\centering
\caption{Architectural configurations of the six surrogate models.}
\label{tab:model_architectures}
\small
\begin{tabular}{>{\raggedright\arraybackslash}p{0.17\textwidth}
                >{\raggedright\arraybackslash}p{0.75\textwidth}}
\toprule
Model & Architectural configuration \\
\midrule
\multicolumn{2}{l}{\textit{Geometry mapping}} \\
Geo-FNO-A
& Residual FNO; 4 Fourier blocks; feature width 36; 16 axial and 8 radial
  Fourier modes; padding $(8,4)$; group normalization; GELU activation;
  trainable parameters: 1.334 M. \\
Geo-FNO-L
& Geo-FNO with learned IPHI mapping; latent grid $130\times40$; 2 geometric
  and 3 standard Fourier layers; feature width 32; 16 axial and 8 radial
  Fourier modes; GELU activation; trainable parameters: 1.349 M. \\
\multicolumn{2}{l}{\textit{Extended-domain representation}} \\
eDAFNO
& Uniform grid: $250\times50$; lifting input $[\boldsymbol{\mu},r,z]$;
  $m_{\beta}$ incorporated into every eDAFNO block; $\beta=100$; 4 explicit
  eDAFNO blocks; latent width 32; 16 axial and 8 radial Fourier modes; padding
  $(48,4)$; GELU activation; trainable parameters: 1.054 M. \\
\multicolumn{2}{l}{\textit{Coordinate-based continuous representation}} \\
C-MLP
& Pointwise input $[\boldsymbol{\mu},r,z]$; 4 hidden layers; width 256; GELU
  activation; trainable parameters: 0.200 M. \\
\multicolumn{2}{l}{\textit{Geometry-native discretizations}} \\
PCNO
& 4 PCNO layers; latent width 64; 4 radial and 4 axial Fourier modes;
  one-hop local least-squares gradient reconstruction; projection width 128;
  GELU activation; trainable parameters: 1.386 M. \\
Transolver
& 6 Physics-Attention blocks; latent width 64; 4 attention heads; head
  dimension 16; 16 physics slices; projection width 128; GELU feed-forward
  layers; zero dropout; trainable parameters: 0.191 M. \\
\bottomrule
\end{tabular}
\end{table}
\section{Sensitivities to Geometry Representation}
\label{app:geometry_sensitivities}
\setcounter{table}{0}
\renewcommand{\thetable}{B\arabic{table}}

To clarify how geometric information enters and constrains surrogate learning,
three controlled studies examine prescribed geometry alignment, explicit
electrode-gap input, and spatial resolution. Unless otherwise stated, only the
factor under investigation is changed, and each variant is trained at
$N=1000$ using the same data splits, loss functions, training settings, and
five paired random seeds as its baseline. All errors are evaluated on the
original solver grids using the same full-domain and wafer-adjacent metrics
defined above.

\subsection{Introducing Prescribed Geometry Alignment}

The original learning problem contains three sources of complexity:
\begin{enumerate}
    \item the four parameters $(p,V,f,g)$ jointly determine the coupled plasma response;
    \item changes in the electrode gap produce different computational domains;
    \item the resulting spatial discretizations lack pointwise correspondence
    across samples.
\end{enumerate}
The prescribed mapping used by Geo-FNO-A removes the latter two sources by
placing every case on a common reference domain with aligned discretization
points. The mapped models therefore need only learn the four-parameter
dependence of the plasma response. For eDAFNO, C-MLP, PCNO, and Transolver,
the training and test datasets are preprocessed onto the common reference
domain using the prescribed mapping before being supplied to the models.
Table~\ref{tab:unified_mapping} compares their original and mapped
representations.

\begin{table}[H]
    \centering
    \caption{Effect of applying the prescribed geometry mapping.}
    \label{tab:unified_mapping}
    \small
    \begin{tabular}{lcccc}
        \toprule
        Model & \multicolumn{2}{c}{Four-field error (\%)}
        & \multicolumn{2}{c}{Wafer-adjacent $n_i$ error (\%)} \\
        \cmidrule(lr){2-3}\cmidrule(lr){4-5}
        & \shortstack{Original$\rightarrow$\\Prescribed mapping} & Change
        & \shortstack{Original$\rightarrow$\\Prescribed mapping} & Change \\
        \midrule
        eDAFNO & $\mathbf{1.32 \rightarrow 0.97}$ & $\mathbf{-26.7\%}$
               & $\mathbf{13.80 \rightarrow 9.23}$ & $\mathbf{-33.1\%}$ \\
        C-MLP & $1.64 \rightarrow 1.88$ & $+14.1\%$
              & $15.29 \rightarrow 16.45$ & $+7.6\%$ \\
        PCNO  & $0.90 \rightarrow 0.92$ & $+1.8\%$
              & $8.00 \rightarrow 8.82$ & $+10.3\%$ \\
        Transolver & $0.85 \rightarrow 0.94$ & $+10.1\%$
                   & $2.76 \rightarrow 3.15$ & $+14.1\%$ \\
        \bottomrule
    \end{tabular}
\end{table}

Table~\ref{tab:unified_mapping} shows that prescribed alignment is beneficial
only when it compensates for a limitation of the original representation. For
eDAFNO, the full-domain and wafer-adjacent errors decrease by $26.7\%$ and
$33.1\%$, respectively. In its original extended-domain representation, the
distance field identifies the boundary location, but the model must still
learn how corresponding plasma structures shift as the electrode gap changes.
The prescribed mapping establishes this correspondence before learning and
therefore better matches the grid-based operator. By contrast, C-MLP, PCNO,
and Transolver already describe each sample through its spatial coordinates or
geometry-native discretizations. Mapping provides little additional geometric
information to these models while introducing interpolation and preprocessing
errors, resulting in unchanged or moderately increased prediction errors.
These results indicate that prescribed alignment is most useful when the
original representation does not directly establish correspondence across
variable domains.

\subsection{Removing the Explicit Electrode-Gap Input}

To determine whether each representation can distinguish variable geometries
without an explicit geometric parameter, the electrode gap $g$ is removed
from the model input. This test examines whether the geometry remains
implicitly encoded in the spatial representation.

\begin{table}[H]
    \centering
    \caption{Effect of removing the explicit electrode-gap input.}
    \label{tab:no_gap_input}
    \small
    \begin{tabular}{lcccc}
        \toprule
        Model & \multicolumn{2}{c}{Four-field error (\%)}
        & \multicolumn{2}{c}{Wafer-adjacent $n_i$ error (\%)} \\
        \cmidrule(lr){2-3}\cmidrule(lr){4-5}
        & \shortstack{Original$\rightarrow$\\Without gap input} & Change
        & \shortstack{Original$\rightarrow$\\Without gap input} & Change \\
        \midrule
        Geo-FNO-A & $\mathbf{0.90 \rightarrow 7.31}$ & $\mathbf{+708.2\%}$
                  & $\mathbf{3.94 \rightarrow 8.92}$ & $\mathbf{+126.5\%}$ \\
        Geo-FNO-L & $1.01 \rightarrow 1.01$ & $-0.1\%$
                  & $7.15 \rightarrow 6.16$ & $-13.9\%$ \\
        eDAFNO & $1.32 \rightarrow 1.31$ & $-1.0\%$
               & $13.80 \rightarrow 13.07$ & $-5.3\%$ \\
        C-MLP & $\mathbf{1.64 \rightarrow 23.35}$ & $\mathbf{+1320.3\%}$
              & $\mathbf{15.29 \rightarrow 431.66}$ & $\mathbf{+2723.6\%}$ \\
        PCNO & $0.90 \rightarrow 0.92$ & $+2.6\%$
             & $8.00 \rightarrow 7.69$ & $-3.9\%$ \\
        Transolver & $0.85 \rightarrow 0.85$ & $+0.5\%$
                   & $2.76 \rightarrow 3.03$ & $+9.7\%$ \\
        \bottomrule
    \end{tabular}
\end{table}

Table~\ref{tab:no_gap_input} reveals two distinct responses. Removing $g$
increases the full-domain errors of Geo-FNO-A and C-MLP from $0.90\%$ to
$7.31\%$ and from $1.64\%$ to $23.35\%$, respectively, whereas the other four
models retain comparable full-domain accuracy. For Geo-FNO-A, the prescribed
mapping has already removed the domain variation and aligned the spatial
coordinates; without $g$, cases with identical $(p,V,f)$ but different gaps
become indistinguishable at the input. C-MLP receives the coordinate of each
query point, but its pointwise predictions cannot infer the global extent of
the computational domain. By contrast, eDAFNO retains the boundary location
through its distance field, while Geo-FNO-L, PCNO, and Transolver couple
information across the spatial discretization and can therefore recover the
geometry from its spatial organization. Explicit gap information is thus
necessary when the representation removes, or the architecture cannot
aggregate, the global domain geometry.

\subsection{Effects of Spatial Resolution}

To examine the sensitivity of each representation to spatial resolution,
reduced-resolution variants are constructed for all six models. The reference
grids of Geo-FNO-A and Geo-FNO-L are reduced from $130\times40$ to
$65\times20$, while the background grid of eDAFNO is reduced from
$250\times50$ to $125\times25$. For C-MLP, PCNO, and Transolver, the native
training point sets are uniformly subsampled in both spatial directions with
boundary points preserved, retaining approximately $26.7\%$ of the original
points. All six models are retrained using these reduced spatial
representations, with predictions evaluated on the original solver grids.

\begin{table}[H]
    \centering
    \caption{Effect of reducing the spatial resolution.}
    \label{tab:low_resolution}
    \small
    \begin{tabular}{lcccc}
        \toprule
        Model & \multicolumn{2}{c}{Four-field error (\%)}
        & \multicolumn{2}{c}{Wafer-adjacent $n_i$ error (\%)} \\
        \cmidrule(lr){2-3}\cmidrule(lr){4-5}
        & \shortstack{Original$\rightarrow$\\Reduced resolution} & Change
        & \shortstack{Original$\rightarrow$\\Reduced resolution} & Change \\
        \midrule
        Geo-FNO-A & $0.90 \rightarrow 1.83$ & $+102.0\%$
                  & $3.94 \rightarrow 5.49$ & $+39.3\%$ \\
        Geo-FNO-L & $1.01 \rightarrow 1.08$ & $+7.4\%$
                  & $7.15 \rightarrow 6.45$ & $-9.7\%$ \\
        eDAFNO & $1.32 \rightarrow 2.12$ & $+60.9\%$
               & $13.80 \rightarrow 21.87$ & $+58.4\%$ \\
        C-MLP & $1.64 \rightarrow 1.64$ & $0.0\%$
              & $15.29 \rightarrow 12.39$ & $-19.0\%$ \\
        PCNO & $0.90 \rightarrow 10.05$ & $+1018.2\%$
             & $8.00 \rightarrow 47.91$ & $+498.7\%$ \\
        Transolver & $0.85 \rightarrow 3.34$ & $+292.3\%$
                   & $2.76 \rightarrow 5.49$ & $+98.8\%$ \\
        \bottomrule
    \end{tabular}
\end{table}

Table~\ref{tab:low_resolution} shows that sensitivity to spatial resolution
depends strongly on how the discretization is used by the architecture.
Geo-FNO-L and C-MLP remain comparatively robust in both the full-domain and
wafer-adjacent metrics. Their continuous coordinate-conditioned predictions
decouple the output locations from the spatial sampling used during training.
Geo-FNO-A and eDAFNO show greater sensitivity because they first represent the
fields on reduced structured grids, where fine-scale and boundary information
may be lost before the predictions are restored to the original grids. PCNO
and Transolver exhibit the largest degradation. For these models, the sampled
points determine not only where the fields are represented but also how
spatial interactions or token groupings are constructed; reducing the
resolution therefore alters the operation of the learned operator itself.
Geometry-native operation consequently does not guarantee robustness to
spatial resolution.

\section{Sensitivities to Surrogate Architecture}
\label{app:architecture_sensitivities}
\setcounter{figure}{0}
\renewcommand{\thefigure}{C\arabic{figure}}

\subsection{Learned Slicing and Token Attention in Transolver}

Transolver shows one of the strongest improvements as the training-set size
increases. To identify the architectural components responsible for this
behavior, two controlled variants are considered. The first replaces the
learned slicing with a fixed $4\times4$ partition while retaining token
attention. The second retains the learned slicing but removes the interaction
among tokens. All other model and training settings remain unchanged, and both
variants are evaluated using five paired seeds at $N=100$ and $N=1000$.

\begin{figure}[!htbp]
    \centering
    \includegraphics[width=0.98\textwidth]{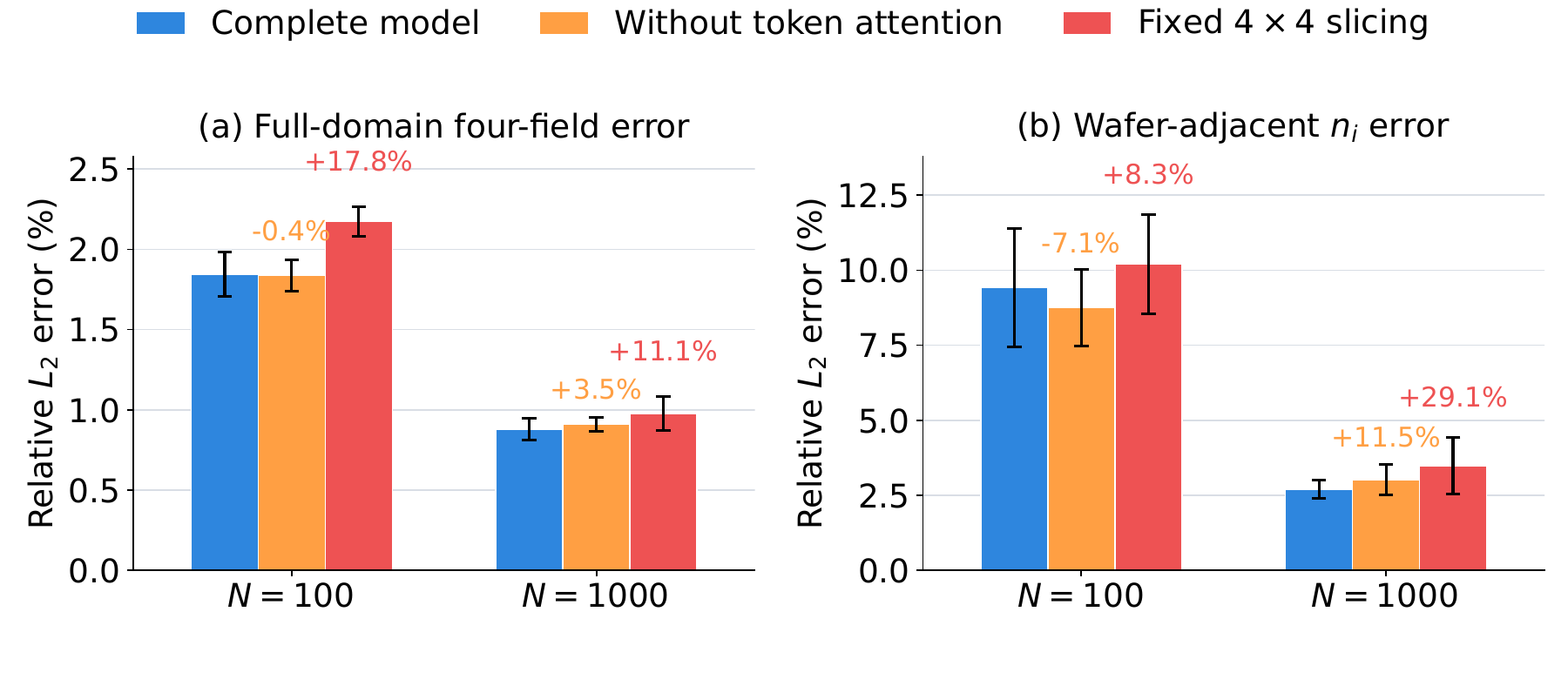}
    \caption{Effects of learned slicing and token attention in Transolver at
    $N=100$ and $N=1000$. The complete model is compared with two variants:
    one replaces learned slicing with a fixed $4\times4$ partition, and the
    other removes token attention. (a) Full-domain four-field error and (b)
    wafer-adjacent ion-density error. Percentages denote the relative error
    changes from the complete model at each training-set size.}
    \label{fig:transolver_structural_ablation}
\end{figure}

Figure~\ref{fig:transolver_structural_ablation} shows that replacing the
learned slicing with a fixed partition increases the four-field error by
$17.8\%$ at $N=100$ and $11.1\%$ at $N=1000$, with a $29.1\%$ increase in the
wafer-adjacent $n_i$ error at $N=1000$. These consistent penalties identify
learned slicing as the main contributor, allowing Transolver to organize the
discretization points according to the plasma state rather than a prescribed
partition. Removing token attention causes little change at $N=100$, but
increases the four-field and wafer-adjacent errors by $3.5\%$ and $11.5\%$,
respectively, at $N=1000$. The benefit of token interaction therefore emerges
only with sufficient training data, helping explain the rapid improvement of
Transolver as the data set grows.

\subsection{Fourier Modes and Domain Padding in Geo-FNO-A}

To examine the roles of spectral resolution and boundary treatment in
Geo-FNO-A, two controlled comparisons are performed at $N=1000$. The complete
model uses $16\times8$ retained Fourier modes and $8\times4$ axial and radial
padding. The first comparison varies the retained modes among $8\times4$,
$16\times8$, and $24\times12$, with the latent width adjusted to maintain a
comparable model capacity. The second varies the padding among $8\times4$,
$8\times0$, and $0\times0$ while retaining the Fourier modes and latent width
of the complete model. All other training settings remain unchanged.

\begin{figure}[!htbp]
    \centering
    \includegraphics[width=0.98\textwidth]{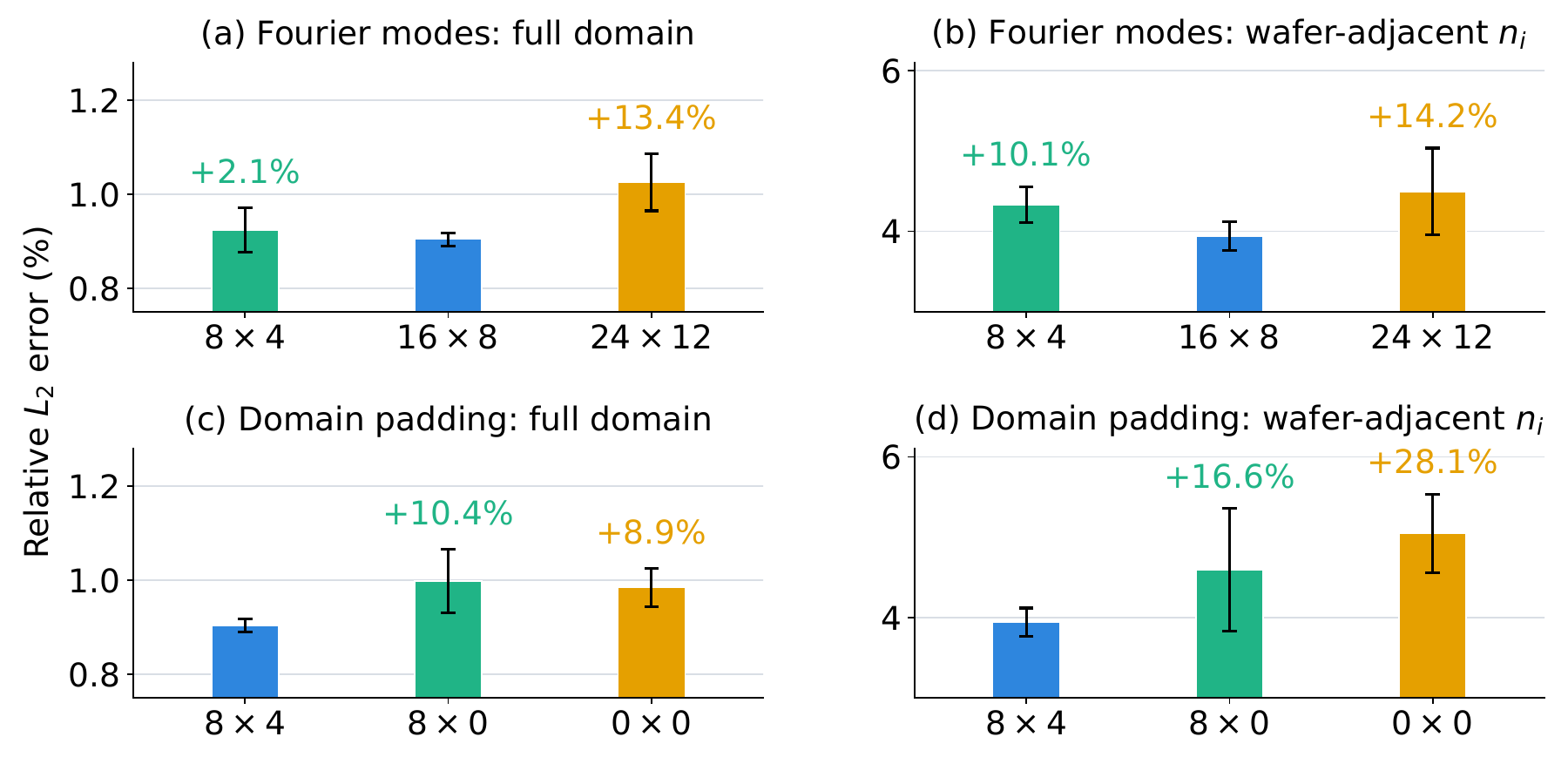}
    \caption{Effects of retained Fourier modes and domain padding in Geo-FNO-A
    at $N=1000$. The complete configuration uses $16\times8$ modes and
    $8\times4$ axial and radial padding. Panels (a,b) compare $8\times4$,
    $16\times8$, and $24\times12$ modes with approximately matched parameter
    counts. Panels (c,d) compare $8\times4$, $8\times0$, and $0\times0$ padding
    with fixed retained modes. The left and right columns report the full-domain
    four-field and wafer-adjacent ion-density errors, respectively. Percentages
    indicate changes relative to the complete model.}
    \label{fig:geofno_mode_padding_ablation}
\end{figure}

Figure~\ref{fig:geofno_mode_padding_ablation} shows that the retained modes
affect global and local accuracy differently. Reducing the modes to
$8\times4$ changes the full-domain error by only $2.1\%$, but increases the
wafer-adjacent $n_i$ error by $10.1\%$. Increasing the modes to $24\times12$
raises the two errors by $13.4\%$ and $14.2\%$, respectively. This
nonmonotonic response indicates that performance depends on the balance
between spectral resolution and latent feature capacity, rather than on the
number of retained modes alone. The padding comparison shows a clearer trend.
Removing radial padding increases the full-domain and wafer-adjacent errors by
$10.4\%$ and $16.6\%$, while removing both axial and radial padding increases
them by $8.9\%$ and $28.1\%$. The larger local penalty shows that full-domain
averaging understates the sensitivity of the wafer-adjacent plasma structure
to spectral boundary treatment. Prescribed mapping therefore aligns the
variable geometries, but domain padding remains necessary to represent the
nonperiodic boundaries of the aligned fields.

\subsection{Operator Branches and Quadrature in PCNO}

To assess the roles of the spatial operators and numerical
quadrature in PCNO, the complete model is compared with three retrained
variants. Two variants disable the local gradient branch or the Fourier
integral branch, respectively. The third replaces the physical-volume
quadrature weights within the Fourier integral operator with equal weights
over the active points. Physical cell-volume weighting is retained in the
training loss and evaluation metrics. The remaining architecture and output
parameterization are retained. All comparisons use $N=1000$, the original training loss,
five paired random seeds, and the same 100 test cases.

\begin{figure}[!htbp]
    \centering
    \includegraphics[width=0.98\textwidth]{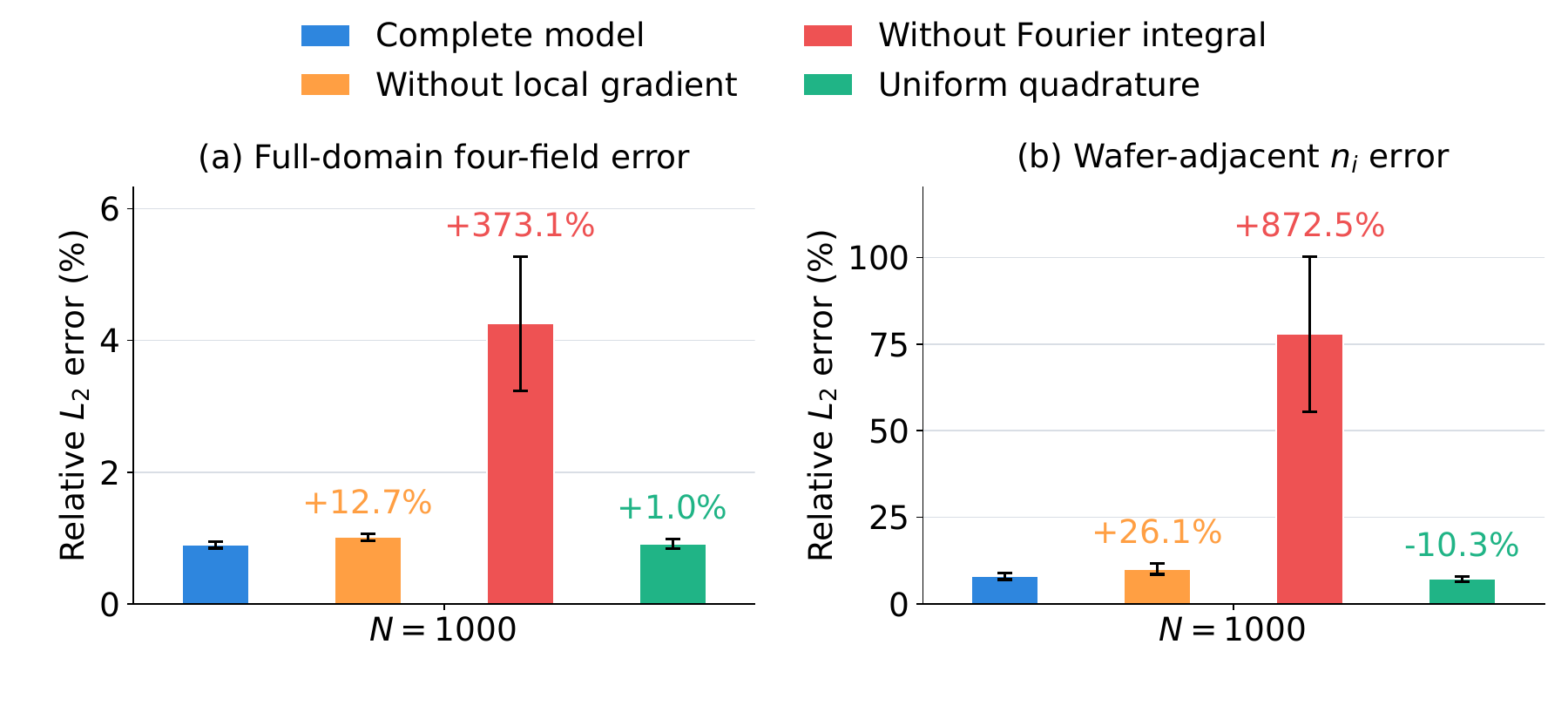}
    \caption{Effects of operator branches and Fourier
    quadrature in PCNO at $N=1000$. The complete model is compared with
    variants trained without the local gradient branch, without the Fourier
    integral branch, or with uniform quadrature weights in the Fourier
    integral operator. Panels (a) and (b) report
    the full-domain four-field and wafer-adjacent ion-density profile
    errors, respectively. Percentages indicate
    relative changes from the complete model's mean error.}
    \label{fig:pcno_operator_ablation}
\end{figure}

Figure~\ref{fig:pcno_operator_ablation} shows that removing
the local gradient branch increases the mean whole-field error from
$0.899\%$ to $1.013\%$ and the wafer-adjacent profile error from $8.003\%$
to $10.093\%$. Removing the Fourier integral branch causes a much larger
increase, to $4.254\%$ and $77.830\%$, respectively. The Fourier integral
branch therefore makes the larger contribution to prediction accuracy in
this benchmark, while the local gradient branch provides an additional
benefit, particularly for the near-wafer profile.
Uniform quadrature gives a mean whole-field error of
$0.909\%$, close to the complete model's $0.899\%$, and reduces the mean
wafer-adjacent profile error from $8.003\%$ to $7.175\%$. Thus, the
physical-volume quadrature weights do not provide a consistent accuracy
advantage over equal weights in the present benchmark.

\bibliographystyle{unsrt}
\bibliography{references}

\end{document}